\documentclass[11pt]{article}

\usepackage[preprint]{acl}

\usepackage{times}
\usepackage{latexsym}

\usepackage[T1]{fontenc}

\usepackage[utf8]{inputenc}

\usepackage{microtype}
\usepackage{amssymb}
\usepackage{pifont}

\usepackage{inconsolata}

\usepackage{tcolorbox}
\usepackage{graphicx}

\usepackage{amsmath}
\usepackage{enumitem}
\usepackage{booktabs}
\usepackage{multicol,multirow}
\usepackage{adjustbox}
\usepackage{makecell}
\usepackage{longtable}

\usepackage{marvosym}
\tcbuselibrary{skins,breakable} 

\newcommand{\ourmethod}{\textsc{P2P}}
\title{Instant Personalized Large Language Model Adaptation via Hypernetwork}

\author{Zhaoxuan Tan$^{\text{\Letter}1}$\thanks{Work done while interning at Amazon.}, Zixuan Zhang$^{2}$, Haoyang Wen$^{2}$, Zheng Li$^{2}$, Rongzhi Zhang$^{2}$, Pei Chen$^{2}$, \\
\textbf{Fengran Mo$^{3*}$, Zheyuan Liu$^{1*}$, Qingkai Zeng$^{1}$\thanks{Corresponding author: Qingkai Zeng, qzeng@nd.edu}, Qingyu Yin$^{2}$, Meng Jiang$^{1}$} \\
$^{1}$University of Notre Dame, \ \ \ $^{2}$Amazon.com Inc, \ \ \ $^{3}$Université de Montréal\\
\texttt{ztan3@nd.edu} \\
}

\newtcolorbox{prompt}[1]{
    enhanced,
    drop shadow=black!5!white,
    left=4mm,
    right=4mm,
    top=2mm,
    bottom=2mm,
    boxsep=0mm,
    rounded corners,
    title=#1,    fontupper=\footnotesize\linespread{0.9}\fontfamily{lmr}\selectfont,
    }

\begin{document}
\maketitle

\begin{abstract}
Personalized large language models (LLMs) tailor content to individual preferences using user profiles or histories. However, existing parameter-efficient fine-tuning (PEFT) methods, such as the ``One-PEFT-Per-User'' (OPPU) paradigm, require training a separate adapter for each user, making them computationally expensive and impractical for real-time updates. We introduce Profile-to-PEFT, a scalable framework that employs a hypernetwork, trained end-to-end, to map a user's encoded profile directly to a full set of adapter parameters (\emph{e.g.}, LoRA), eliminating per-user training at deployment. This design enables instant adaptation, generalization to unseen users, and privacy-preserving local deployment. Experimental results demonstrate that our method outperforms both prompt-based personalization and OPPU while using substantially fewer computational resources at deployment. The framework exhibits strong generalization to out-of-distribution users and maintains robustness across varying user activity levels and different embedding backbones. The proposed Profile-to-PEFT framework enables efficient, scalable, and adaptive LLM personalization suitable for large-scale applications. Our implementation is available at \url{https://zhaoxuan.info/p2p.github.io/}.
\end{abstract}

\section{Introduction}
Personalization aims to tailor system interactions, content, and recommendations to a user's specific needs and preferences by leveraging their historical data \citep{tan2023user, chen2023large, kirk2024benefits,liu2025survey}. While large language models (LLMs) have demonstrated powerful generative capabilities, their general-purpose, ``one-size-fits-all" nature limits their ability to cater to individual users \citep{guan-etal-2025-survey, Zhang2024PersonalizationOL}. Consequently, integrating the generative strength of LLMs with user-specific personalization has become a critical research direction \citep{li2023teach, jiang2025personamem, DBLP:conf/acl/Tan0LWY0C000NY025}.

\begin{figure}[t]
    \centering
    \includegraphics[width=1\linewidth]{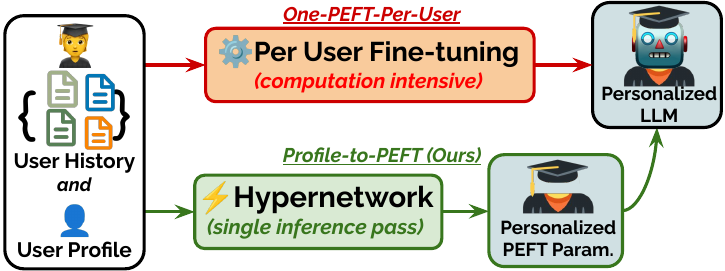}
    \caption{The "One-PEFT-Per-User" method uses computationally intensive fine-tuning to create personalized parameters. In contrast, our proposed Profile-to-PEFT uses a hypernetwork to directly generate parameters from user history or profile in a single inference pass.}
    \vspace*{-10pt}
    \label{fig:teaser}
\end{figure}

Recent methodologies fall into two main categories: prompt-based and parameter-efficient fine-tuning (PEFT)-based. Prompt-based personalization techniques design specific prompt templates to guide LLMs in capturing user preferences, encompassing approaches such as vanilla personalized prompting \citep{zhiyuli2023bookgpt}, retrieval-augmented prompting \citep{salemi2023lamp}, and profile-augmented prompting \citep{richardson2023integrating}. However, these methods expose user data to centralized LLMs, raising significant concerns regarding user privacy and hindering model ownership for deep personalization \citep{tan2024democratizing}. Additionally, prompt-based techniques are susceptible to distraction by irrelevant user historical data, an issue difficult to mitigate solely through incorporating additional retrieval context \citep{shi2023large}. Conversely, PEFT-based personalization strategies store users’ preferences and behavioral patterns in lightweight, user-specific parameters. OPPU \citep{tan2024democratizing} is a pioneering PEFT-based approach that effectively encodes user preferences into individual PEFT parameters, facilitating model ownership, stronger personalization performance, and superior generalization of user behavior patterns compared to prompt-based methods.

Despite their effectiveness, existing PEFT-based frameworks operate under a "one-PEFT-per-user" paradigm, where a unique module is trained from scratch for each user. This approach presents substantial computational and scalability challenges, particularly in large-scale systems with millions of users or in dynamic settings where user preferences evolve continuously. Training or updating individual PEFT modules in real-time is computationally prohibitive. This bottleneck leads to a key research question: \textit{Is it possible to generate personalized PEFT parameters directly from a user's profile in an efficient step, thereby eliminating the need for per-user training at deployment?}

To address this challenge, we introduce Profile-to-PEFT (\ourmethod{}), a novel framework that learns a direct mapping from user profiles to personalized PEFT parameters. Instead of relying on iterative fine-tuning, our method utilizes a hypernetwork that generates a full set of personalized LoRA adapter weights conditioned on a user's profile, thereby eliminating the need to perform per user training at deployment, as illustrated in Figure \ref{fig:teaser}. The process, detailed in Figure \ref{fig:overview}, begins by constructing the user profile composed of natural language user preference summaries from user history and retrieved historical interactions, then compact the profile into user embeddings. This embedding, augmented with learnable position and module identifiers, is then fed into an MLP-based hypernetwork, which outputs the entire set of personalized adapter parameters in a single forward pass. By plugging in the user-specific parameters and train this framework end-to-end on a diverse user population data using supervised finetuning, \ourmethod{} learns to generalize across unseen users. 

We conduct extensive experiments on LaMP \cite{salemi2023lamp}, LongLaMP \cite{kumar2024longlamp}, Personal Reddit \cite{staab2023beyond}, and Empathic Conversations \cite{omitaomu2022empathic} datasets that containing diverse classification and generation tasks. The results, corroborated by LLM-as-a-Judge evaluations, demonstrate that \ourmethod{} generalizes effectively to both in-distribution and out-of-distribution users. Our analyses confirm that training user diversity is more critical than sheer quantity for robust performance and that our generation-based approach is 33x faster at deployment than the OPPU paradigm. Further studies validate the framework's robustness to different embedding models and varying user activity levels, confirming our key design choices.


In summary, the proposed \ourmethod{} framework advances PEFT-based personalized LLM towards practical deployment at industrial scales, enabling strong generalization to unseen users during training, and achieve real-time personalized adaptations of LLMs. \ourmethod{} maintains user privacy and significantly reduces the computational burden and carbon footprint of personalized LLM training.

\section{Preliminary}
\paragraph{Research Problem Formulation} 
We aim to personalize LLMs for individual users. At time $t$, a user $u$ with history $\mathcal{H}_u^{t}$ (containing all behaviors before $t$) queries the model with input $x_u$ to receive personalized output $y_u$. The goal is to obtain personalized parameters $\Delta W_u$ for each user $u$ or $\Delta W_{x_u}$ for each input $x_u$ of user $u$.

\paragraph{Low-Rank Adaptation (LoRA)} \cite{hu2021lora} is a PEFT method that freezes pre-trained weights of a LLM $W_0 \in \mathbb{R}^{d_{out} \times d_{in}}$ and introduce trainable low-rank matrices $\Delta W = BA$, where $B \in \mathbb{R}^{d_{out} \times r}$ and $A \in \mathbb{R}^{r \times d_{in}}$, with the rank $r \ll \min(d_{in}, d_{out})$. The model's forward pass becomes $h = W_0x + \Delta W x = W_0x + BAx$. We denote LoRA weights for module $m$ at layer $l$ as $\Delta W^{m,l}$.


\paragraph{Personalization via Per-User PEFT} \cite{tan2024democratizing}
trains unique PEFT parameters for each user by using the following objective:
\begin{align*}
    \Delta W_u^* = \arg\min_{\Delta W} \mathcal{L}_{\text{SFT}}(\Psi \oplus \Delta W, \mathcal{H}_u^{<t}),
\end{align*}
where $\Psi$ denotes frozen base model weights, $\mathcal{L}_{\mathrm{SFT}}$ is the supervised fine-tuning loss, and $\oplus$ applies PEFT to the base model. While effective, this requires separate training for every user, limiting scalability and real-time adaptation.


\section{Profile-to-PEFT (\ourmethod{})}
\begin{figure}[t]
    \centering
    \includegraphics[width=1\linewidth]{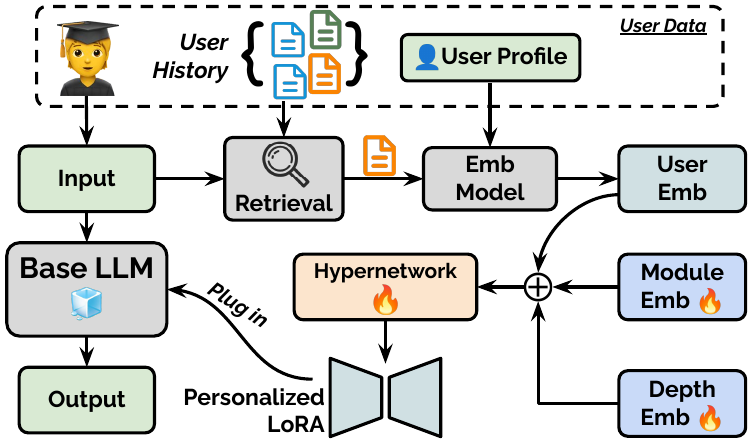}
    \caption{Overview of the Profile-to-PEFT architecture, where user history, depth, module embeddings are fed into the hypernetwork to obtain personalized LoRA. \ourmethod{} is optimized in a end-to-end training manner.}
    \label{fig:overview}
\end{figure}

To address the limitations of the one-PEFT-per-user paradigm, we present Profile-to-PEFT (\ourmethod{}), which learns a direct mapping from a user profile to PEFT parameters. Instead of running iterative per-user optimization, \ourmethod{} uses a hypernetwork $f_\theta$ to produce personalized LoRA weights in a single forward pass, as illustrated in Figure \ref{fig:overview}.


\subsection{Model Architecture}
\ourmethod{} generates the LoRA matrices ($A_{x_u}^{m,l}, B_{x_u}^{m,l}$) for each target module $m$ at layer $l$ conditioned on a user profile. This process involves three key steps.

\paragraph{User Profile Encoding} 
First, a textual user profile $p_u^{<t}$ is constructed to dynamically represent user preferences. This profile combines a global summary $s_u^{<t}=\mathrm{Profiler}(\mathcal{H}_u^{<t})$, generated from the user's history by the base LLM, with the top-$k$ most relevant historical interactions retrieved by a retriever $\mathcal{R}$ conditioned on the current input $x_u$. The profile text is formulated as:
\begin{align*}
    p_{x_u}^{<t}=\left[s_u^{<t} || \mathcal{R}(x_u, \mathcal{H}_u^{<t}, k)\right].
\end{align*}
If a pre-existing profile is available in the data, we use it directly. This text is then encoded into a fixed-dimensional user embedding $e_u$ using a frozen sentence embedding model $\mathrm{Enc}(\cdot)$, such that $e_{x_u} = \mathrm{Enc}(p_{x_u}^{<t})$. This embedding serves as a condensed representation of the user's preferences and behavioral patterns.

\paragraph{Position-Aware Input Formulation} 
To enable the hypernetwork to generate distinct parameters for different locations within the LLM, the user embedding $e_u$ is augmented with learnable positional embeddings. For a specific module $m$ at layer $l$, the input representation $\phi_{x_u}^{m,l}$ is formed by concatenating $e_{x_u}$ with a module embedding $E_{\text{mod}}[m]$ and a depth embedding $E_{\text{dep}}[l]$:
\begin{align*}
\phi_{x_u}^{m,l} = \big[\, e_{x_u} \, || \, E_{\mathrm{mod}}[m] \, || \, E_{\mathrm{dep}}[l] \,\big].
\end{align*}
\paragraph{Parameter Generation} 
The position-aware representation $\phi_u^{m,l}$ is passed through the hypernetwork $f_\theta$, which is implemented as an MLP. The hypernetwork outputs a flattened parameter vector that is then reshaped and split to form the low-rank LoRA matrices, $A_u^{m,l}$ and $B_u^{m,l}$, expressed as
\begin{align*}
(A_{x_u}^{m,l}, B_{x_u}^{m,l}) = \mathrm{Unflatten}(f_\theta(\phi_{x_u}^{m,l})).
\end{align*}
This process is batched over all target positions $(m,l) \in \mathcal{I}$, where $\mathcal{I}$ is the set of all selected LoRA modules. We denote the complete set of generated parameters for user $u$ as $\Delta W_{x_u} = \mathrm{Gen}_\theta(p_{x_u}^{<t})$.

\subsection{Training and Inference}
We optimize the hypernetwork parameters $\theta$ in an end-to-end fashion across a diverse population of training users. For each user, we use their profile $p_{x_u}^{<t}$ to generate a full set of PEFT parameters. The objective is to minimize the supervised fine-tuning loss on the user's subsequent interactions $\mathcal{H}_u^{\geqslant t}$ when these parameters are applied to the base model $\Psi$. The training objective is formulated as
\begin{align*}
\mathcal{L}(\theta) = \mathbb{E}_{u \sim \mathcal{U}} \left[ \mathcal{L}_{\text{SFT}}\left( \Psi \oplus \mathrm{Gen}_\theta(p_{x_u}^{<t}), \mathcal{H}_u^{\geqslant t} \right) \right].
\end{align*}
By training on a wide variety of users and personalization tasks, the hypernetwork $f_\theta$ learns a generalized mapping from natural language user profile to personalized PEFT parameters.

At deployment, the trained hypernetwork enables highly efficient personalization. For any user, including those unseen during training $u\notin \mathcal{U}$, their profile $p_u$ is passed through the generator $\mathrm{Gen}_\theta$ to produce personalized weights $\Delta W_u$ in a single inference pass. This on-demand generation obviates the need for per-user fine-tuning, facilitating scalable and real-time LLM personalization. Furthermore, this framework enhances user privacy; when deployed locally, the hypernetwork can process on-device user data without transmitting sensitive information to external servers. This results in a personalization system that is efficient, scalable, privacy-preserving, and continuously adaptive.

\section{Experiment Settings}

\begin{table*}[t]
    \caption{Main experiment results on the LaMP and LongLaMP benchmarks under the \textit{Random split} setting. $\uparrow$ indicates that higher values are better, and $\downarrow$ implies lower values are preferable. For each task, the best score is in \textbf{bold} and the second best is \underline{underlined}. The final row reports average per-instance inference time (ms).}
    \centering
    \begin{adjustbox}{max width=0.9\linewidth}
        \begin{tabular}{llcccc|cc|c}
        \toprule[1.5pt]
        \multirow{2}{*}{\textbf{Task}} & \multirow{2}{*}{\textbf{Metric}} & \multirow{2}{*}{\makecell{\textbf{Base}\\\textbf{Model}}} & \multirow{2}{*}{\textbf{RAG}} & \multirow{2}{*}{\textbf{PAG}} & \multirow{2}{*}{\makecell{\textbf{Full}\\\textbf{History}}} & \multirow{2}{*}{\makecell{\textbf{MT}\\\textbf{LoRA}}} & \multirow{2}{*}{\makecell{\textbf{OPPU}}} & \multirow{2}{*}{\textbf{\makecell{\ourmethod{} \\(Ours)}}} \\
        & & & & & & & \\
        \midrule[0.75pt]

        \multirow{2}{*}{\makecell[l]{\textsc{LaMP-1:}\\\textsc{Citation Id.}}} & Acc $\uparrow$ & .519 & .504 & \underline{.563} & .562 & .511 & .531 & \underline{.583}\\
        & F1 $\uparrow$ & .516 & .409 & \underline{.560} & .551 & .507 & .531 & \underline{.580} \\
        \hline
        \multirow{2}{*}{\makecell[l]{\textsc{LaMP-2N: }\\\textsc{News Cat.}}} & Acc $\uparrow$& .653 & .666 & \underline{.761} & .750 & .670 & \textbf{.781} & .716 \\
        & F1 $\uparrow$& .679 & .679 & \underline{.773} & .764 & .683 & \textbf{.782} & .711\\
        \hline
        \multirow{2}{*}{\makecell[l]{\textsc{LaMP-2M: }\\\textsc{Movie Tagging}}} & Acc $\uparrow$& .345 & .351 & .372 & \underline{.413} & .386 & .391 & \textbf{.442}\\
        & F1 $\uparrow$& .292 & .328 & .359 & \underline{.384} & .336 & .359 & \textbf{.408} \\
        \hline
        \multirow{2}{*}{\makecell[l]{\textsc{LaMP-3: }\\\textsc{Product Rating}}} & MAE $\downarrow$& .452 & .553 & .371 & \underline{.344} & .398 & \textbf{.281} & .383\\
        & RMSE $\downarrow$& .801 & 1.00 & .699 & .724 & .731 & \textbf{.617} & \underline{.670}\\
        \hline
        \multirow{2}{*}{\makecell[l]{\textsc{LaMP-4:}\\\textsc{News Headline Gen.}}} & R-1 $\uparrow$& .121 & .130 & .128 & .144 & .150 & \textbf{.167} &.\underline{160}\\
        & R-L $\uparrow$& .108 & .118 & .116 & .131 & .135 & \textbf{.148} & \underline{.145}\\
        \hline
        \multirow{2}{*}{\makecell[l]{\textsc{LaMP-5: }\\\textsc{Scholarly Title Gen.}}} & R-1 $\uparrow$& .465 & \underline{.474} & .436 & .472 & .473 & .468 & \textbf{.490}\\
        & R-L $\uparrow$& .404 & .414 & .382 & .412 & .409 & \underline{.422} & \textbf{.431} \\
        \hline
        \multirow{2}{*}{\makecell[l]{\textsc{LaMP-7:}\\\textsc{Tweet Paraphrase}}} & R-1 $\uparrow$& .376 & .378 & .378 & \underline{.407} & .382 & .353 & \textbf{.442}\\
        & R-L $\uparrow$& .324 & .325 & .326 & \underline{.353} & .332 & .305 & \textbf{.437}\\
        \hline
        \multirow{2}{*}{\makecell[l]{\textsc{LongLaMP-1:}\\\textsc{Abstract Gen.}}} & R-1 $\uparrow$& .267 & .319 & \underline{.331} & .326 & .288 & \textbf{.341} & .314 \\
        & R-L $\uparrow$& .155 & .180 & .181 & \underline{.185} & .165 & \textbf{.195} & .177\\
        \hline
        \multirow{2}{*}{\makecell[l]{\textsc{LongLaMP-2:}\\\textsc{Topic Writing}}} & R-1 $\uparrow$& .283 & .267 & \textbf{.292} & .274 & .248 & .208 & \underline{.284}\\
        & R-L $\uparrow$& .129 & .128 & \underline{.134} & .132 & .122 & .115 & \textbf{.135} \\
        \hline
        \multirow{2}{*}{\makecell[l]{\textsc{LongLaMP-3:}\\\textsc{Review Writing}}} & R-1 $\uparrow$& .212 & .240 & .308 & .235 & .231 & 266 & \textbf{.247}\\
        & R-L $\uparrow$& .121 & .128 & \textbf{.145} & .128 & .120 & \underline{.143} & .137 \\
        \midrule[0.75pt]
        \multicolumn{9}{c}{\textit{Average Performance}} \\
        \midrule[0.75pt]
        \multirow{2}{*}{\makecell[l]{\textsc{Classification} }} & Acc $\uparrow$ & .505 & .507 & .565 & \underline{.575} & .522 & .568 & \textbf{.580}\\
        & F1 $\uparrow$ & .496 & .472 & \underline{.564} & \textbf{.566} & .509 & .557 & \textbf{.566}\\
        \hline
        \multirow{2}{*}{\makecell[l]{\textsc{Generation}}} & R-1 $\uparrow$ & .287 & .301 & \underline{.312} & .310 & .295 & .301 & \textbf{.322}\\
        & R-L $\uparrow$ & .207 & .216 & .214 & \underline{.224} & .214 & .221 & \textbf{.244} \\
        \hline
        \hline
        \textsc{Infer. Time} & ms $\downarrow$ & 31.97 & 44.58 & 66.85 & 461.83 & 30.51 & 35.82 & 39.98 \\
        \bottomrule[1.5pt]
        \end{tabular}%
        \end{adjustbox}
        \label{tab:random_results}
\end{table*}

\paragraph{Baselines} We compare \ourmethod{} against non-personalized base model, retrieval-augmented generation (RAG) \citep{salemi2023lamp}, profile-augmented generation (PAG) \citep{richardson2023integrating}, full user history as context for generation, multi-task LoRA (MT-LoRA), and One PEFT Per User (OPPU) \citep{tan2024democratizing}.\footnote{Please see baseline details in Appendix \ref{app:baseline}.} OPPU directly train personal PEFT on the test user history, which can be envisioned as oracle performance. MT-LoRA is trained without user context, and is considered as task-level adaptation without personalization. For all retrieval operations, we use BM25 \citep{trotman2014improvements} for efficiency and a fair comparison, and set the number of retrieved items to 2 by default. RAG, PAG, and Full History are prompt-based method fine-tuning, while MT LoRA, OPPU, and \ourmethod{} request training, we set $r$ to 8 in LoRA for fair comparison. For all methods, we use \texttt{Qwen2.5-7B-Instruct} \cite{qwen2024qwen2} as the base model and \texttt{Qwen3-Emb-4B} \cite{zhang2025qwen3} as the embedding model. 

\paragraph{Datasets} We employ LaMP \citep{salemi2023lamp}, LongLaMP \citep{kumar2024longlamp}, Personal Reddit (PR) \citep{staab2023beyond}, and Empathetic Conversation (EC) \citep{omitaomu2022empathic} datasets in experiments.\footnote{Please see task details in Appendix \ref{app:tasks}.} LaMP consists of three classification tasks (citation identification, movie tagging, news categorization), one rating prediction task, and three text generation tasks (news headline, scholarly title, tweet paraphrasing) and LongLaMP consists of three long-form generation tasks (abstraction generation, topic writing, review writing). All LaMP and LongLaMP tasks contain per-user behavior history, along with query inputs and ground-truth outputs. Empathetic Conversation consists of essay responses based on the article, Personal Reddit consists of Reddit posts. In PR and EC, each user has a textual profile describing user demographic information and personality traits, as well as user inputs and user-written outputs. 

\paragraph{Evaluation Settings}
To assess generalization, we evaluate our model under two data splits: \textit{random} and \textit{out-of-distribution (OOD)}. For the random split, we sample 200 users from each task as the test set; if a task has fewer than 1000 users, we use a standard 80\%/20\% train-test split. For the OOD split, we construct a challenging test set of users most dissimilar to the training population. To do this, we encode each user's profile into an embedding via \texttt{Qwen3-Emb-4B}, perform kmeans clustering, and select smaller and isolated clusters as test set. The OOD test set size mirrors that of the random split for each task, while all remaining users are used as training set.\footnote{Additional details and statistics are in Appendix~\ref{app:split} and \ref{app:statics}.}

\begin{table*}[t]
    \caption{Main experiment results on the LaMP and LongLaMP benchmarks under \textit{OOD split} setting. $\uparrow$ indicates that higher values are better, and $\downarrow$ implies lower values are preferable. For each task, the best score is in \textbf{bold} and the second best is \underline{underlined}. The final row reports average per-instance inference time (ms).}
    \centering
    \begin{adjustbox}{max width=0.9\linewidth}
        \begin{tabular}{llcccc|cc|c}
        \toprule[1.5pt]
        \multirow{2}{*}{\textbf{Task}} & \multirow{2}{*}{\textbf{Metric}} & \multirow{2}{*}{\makecell{\textbf{Non-}\\\textbf{Personalized}}} & \multirow{2}{*}{\textbf{RAG}} & \multirow{2}{*}{\textbf{PAG}} & \multirow{2}{*}{\makecell{\textbf{Full}\\\textbf{History}}} & \multirow{2}{*}{\makecell{\textbf{MT}\\\textbf{LoRA}}} & \multirow{2}{*}{\textbf{OPPU}} & \multirow{2}{*}{\makecell{\textbf{\ourmethod{}}\\ \textbf{(Ours)}}} \\
        & & & & & & & \\
        \midrule[0.75pt]

        \multirow{2}{*}{\makecell[l]{\textsc{LaMP-1: Personalized}\\\textsc{Citation Identification}}} & Acc $\uparrow$ & .568 & .494 & \textbf{.600} & .592 & .561 & .556 & \underline{.576}\\
        & F1 $\uparrow$ & .569 & .419 & \textbf{.600} & \underline{.584} & .561 & .554 & .577 \\
        \hline
        \multirow{2}{*}{\makecell[l]{\textsc{LaMP-2N: Personalized}\\\textsc{News Categorization}}} & Acc $\uparrow$& .579 & .615 & .623 & \textbf{.655} & .600 & .558 & \underline{.624}\\
        & F1 $\uparrow$&  .601 & \underline{.639} & .630 & \textbf{.665} & .611 & .552 & .612\\
        \hline
        \multirow{2}{*}{\makecell[l]{\textsc{LaMP-2M: Personalized}\\\textsc{Movie Tagging}}} & Acc $\uparrow$& .449 & \underline{.494} & .464 & .479 & .447 & .471 & \textbf{.543}\\
        & F1 $\uparrow$& .406 & \underline{.483} & .461 & .454 & .410 & .416 & \textbf{.497}\\
        \hline
        \multirow{2}{*}{\makecell[l]{\textsc{LaMP-3: Personalized}\\\textsc{Product Rating}}} & MAE $\downarrow$& .465 & .594 & .378 & .290 & .410 & \textbf{.198} & \underline{.258}\\
        & RMSE $\downarrow$& .789 & 1.07 & .700 & .741 & .732 & \textbf{.540} & \underline{.583}\\
        \hline
        \multirow{2}{*}{\makecell[l]{\textsc{LaMP-4: Personalized}\\\textsc{News Headline Gen.}}} & R-1 $\uparrow$& .164 & .184 & .176 & .198 & .184 & \underline{.200} & \textbf{.210}\\
        & R-L $\uparrow$ & .140 & .164 & .154 & .174 & .165 & \underline{.180} & \textbf{.190} \\
        \hline
        \multirow{2}{*}{\makecell[l]{\textsc{LaMP-5: Personalized}\\\textsc{Scholarly Title Gen.}}} & R-1 $\uparrow$& .455 & .473 & .448 & \underline{.495} & .470 & .468 & \textbf{.481} \\
        & R-L $\uparrow$& .401 & .417 & .392 & \textbf{.433} & .426 & .422 & \underline{.431}\\
        \hline
        \multirow{2}{*}{\makecell[l]{\textsc{LaMP-7: Personalized}\\\textsc{Tweet Paraphrasing}}} & R-1 $\uparrow$& .392 & .438 & .449 & \textbf{.479} & .393 & .379 & \underline{.473}\\
        & R-L $\uparrow$& .331 & .385 &.404 & \textbf{.424} & .339 & .323 & \underline{.411}\\
        \hline
        \multirow{2}{*}{\makecell[l]{\textsc{LongLaMP-1:}\\\textsc{Abstract Generation}}} & R-1 $\uparrow$& .267 & .312 & \underline{.323} & .321 & .288 & \textbf{.326} & .301 \\
        & R-L $\uparrow$& .153 & {.177} & .175 & \underline{.179} & .164 & \textbf{.187} & .174\\
        \hline
        \multirow{2}{*}{\makecell[l]{\textsc{LongLaMP-2:}\\\textsc{Topic Writing}}} & R-1 $\uparrow$& .284 & .279 & \textbf{.287} &.280 & .268 & .209 & \underline{.284} \\
        & R-L $\uparrow$& .129 & .135 & \underline{.136} & \textbf{.142} & .130 & .116 & .134\\
        \hline
        \multirow{2}{*}{\makecell[l]{\textsc{LongLaMP-3:}\\\textsc{Review Writing}}} & R-1 $\uparrow$& .204 & .232 & \textbf{.292} & .234 & .170  &\underline{.247} & .220 \\
        & R-L $\uparrow$& .114 & .125 & \textbf{.140} &.124 & .104 & \underline{.130} & .123\\
        \midrule[0.75pt]
        \multicolumn{9}{c}{\textit{Average Performance}} \\
        \midrule[0.75pt]
        \multirow{2}{*}{\makecell[l]{\textsc{Classification} }} & Acc $\uparrow$ & .532 & .534 & .562 & \underline{.575} & .536 & .528 & \textbf{.581} \\
        & F1 $\uparrow$ & .525 & .513 & \underline{.563} & \textbf{.567} & .527 & .507 & \underline{.563}\\
        \hline
        \multirow{2}{*}{\makecell[l]{\textsc{Generation}}} & R-1 $\uparrow$ & .294 & .319 & \underline{.329} & \textbf{.334} & .295 & .305 & .326 \\
        & R-L $\uparrow$ & .211 & .234 & .234 & \textbf{.246} & .221 & .226 & \underline{.243}\\
        \hline
        \hline
        \textsc{Infer. Time} & ms $\downarrow$ & 20.52 & 36.44 & 61.66 & 392.97 & 21.91 & 26.78 & 28.64\\
        \bottomrule[1.5pt]
        \end{tabular}%
        \end{adjustbox}
        \vspace{-0.3cm}
        \label{tab:ood_results}
\end{table*}

\paragraph{Evaluation Metrics}
We use task-appropriate metrics to measure performance. For classification, we report Accuracy and F1-score. For text generation, we use ROUGE-1 and ROUGE-L \citep{lin-2004-rouge}, while for rating prediction, we use RMSE and MAE. For the open-ended generation tasks (Empathetic Conversation, Personal Reddit), we employ a LLM judge (\texttt{GPT-4o}) to assess personalization quality. We adopt the Prometheus prompt \cite{kim2023prometheus} with the user profile and rates the output on a 1-5 scale \emph{w.r.t} the user preferences.

\section{Results}
Tables \ref{tab:random_results} and \ref{tab:ood_results} present our main results on the LaMP and LongLaMP benchmarks, while Table \ref{tab:llm_eval} details the LLM-as-a-Judge evaluations for the PR and EC datasets. The key findings are as follows.

\begin{table}[t]
   \caption{Average LLM-as-a-Judge evaluation scores (on a 1-5 scale) for the Personal Reddit and Empathetic Conversation datasets, comparing performance on \textit{Random} and \textit{OOD} test splits.}
   \centering
   \small
   \begin{adjustbox}{max width=1\linewidth}
   \begin{tabular}{lcccc}
   \toprule[1.2pt]
   \multirow{2}{*}{\textbf{Method}} & \multicolumn{2}{c}{\textbf{Personal Reddit}} & \multicolumn{2}{c}{\textbf{Empathetic Conv.}} \\
   \cmidrule(lr){2-3} \cmidrule(lr){4-5}
   & \textit{Random} & \textit{OOD} & \textit{Random} & \textit{OOD}  \\
   \midrule[0.75pt]
   \textsc{Base Model} & 1.71 & 1.58  &1.86 & \underline{1.55} \\
   \textsc{PAG} & 1.77 & 1.60 & \underline{1.80} & 1.54\\
   \textsc{MT-LoRA}& \underline{1.98} & \underline{1.96} & 1.62 & 1.43\\
   \ourmethod{} (Ours) & \textbf{2.21} & \textbf{2.15} & \textbf{2.03} & \textbf{1.65}\\ 
   \bottomrule[1.2pt]
   \end{tabular}
   \end{adjustbox}
   \vspace{-0.3cm}
   \label{tab:llm_eval}
\end{table}

\begin{figure*}[t]
    \vspace{-0.3cm}
    \centering
    \includegraphics[width=0.95\linewidth]{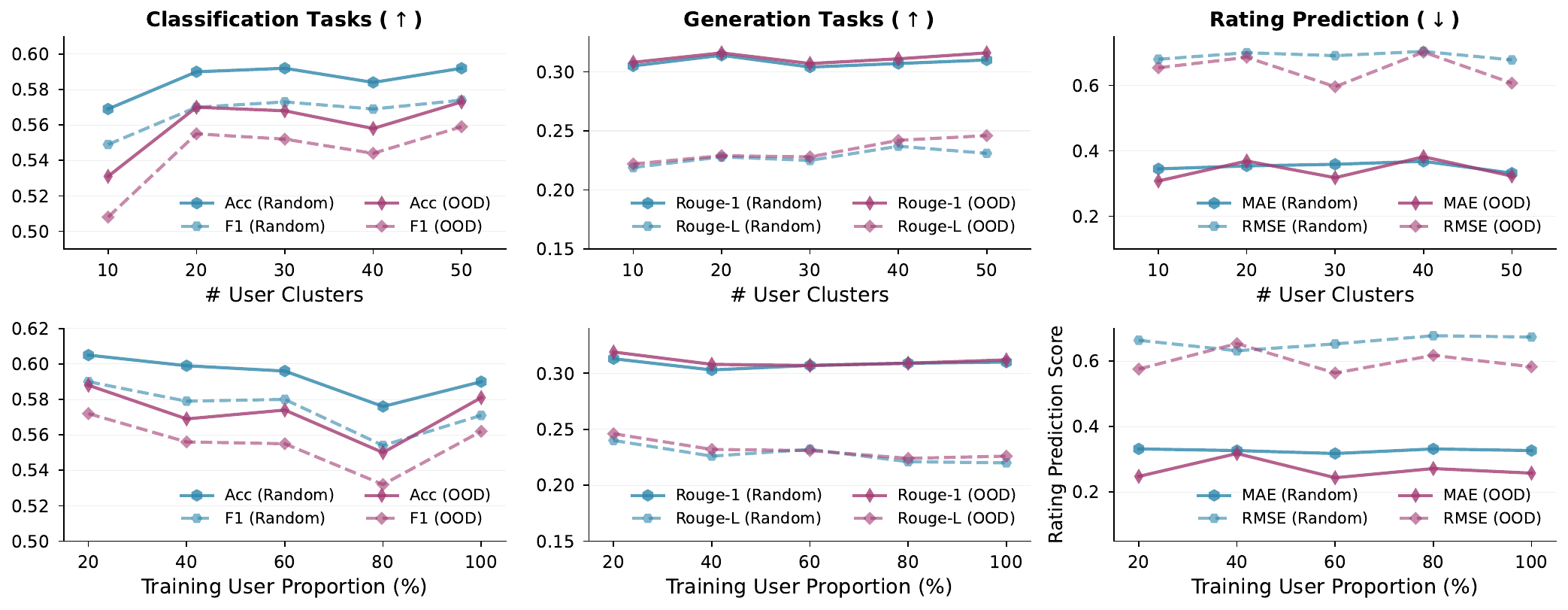}
    \caption{Performance of \ourmethod{} as a function of training user diversity (top row) and quantity (bottom row). While greater diversity boosts performance, increasing the number of users yields no significant gains.}
    \vspace{-0.3cm}
    \label{fig:scaling_users}
\end{figure*}

\paragraph{\ourmethod{} outperforms prompt-based and PEFT-based baselines.}
Across the majority of tasks in the random split setting (Table \ref{tab:random_results}), \ourmethod{} demonstrates superior or highly competitive performance. On average, it achieves the highest accuracy in classification tasks (0.577) and the best ROUGE-L scores in generation tasks (0.244). For instance, in the Tweet Paraphrasing task, \ourmethod{} achieves an ROUGE-1 score of 0.442, significantly outperforming the next best method Full History (0.407). Compared to the PEFT-based OPPU baseline, which requires expensive per-user training, \ourmethod{} achieves better average performance in both classification and generation without any user-specific fine-tuning at deployment. This highlights its ability to effectively generate high-quality personalized parameters in a single forward pass.

\paragraph{\ourmethod{} generalizes well to OOD users.}
The framework demonstrates strong generalization to out-of-distribution users (Table \ref{tab:ood_results}). In this challenging split, \ourmethod{} consistently outperforms other parameter-based baselines like MT-LoRA and OPPU. It also achieves competitive performance against strong prompt-based methods such as PAG and Full History, but with the critical advantage of significantly faster inference and better user privacy preservation. By encoding user history into compact PEFT parameters rather than placing it in the prompt, \ourmethod{} avoids the substantial computational overhead of processing long contexts with every query and exposing user data to the centralized LLM. The outperformance over OPPU is particularly noteworthy because OPPU is fine-tuned directly on the target user's history, whereas \ourmethod{} generates parameters without any user-specific training. This suggests that \ourmethod{} effectively distills collaborative knowledge from the diverse training population, indicating that learning a generalizable mapping from profiles to personalized parameters is a more robust and efficient strategy.

\paragraph{\ourmethod{} excels at open-ended personalized generation.}
For tasks requiring nuanced, open-ended generation, LLM-as-a-Judge evaluations (Table \ref{tab:llm_eval}) confirm the effectiveness of \ourmethod{}. On both the Personal Reddit and Empathetic Conversation datasets, \ourmethod{} consistently achieves the highest scores from the \texttt{GPT-4o} judge, surpassing both task-level MT-LoRA and prompt-based PAG baselines. Interestingly, combining task-level MT-LoRA adaptation with a RAG approach (appending the user profile to the prompt) does not improve performance under this evaluation paradigm, indicating that these methods struggle to generalize for personalized preference adaptation in open-ended scenarios. Moreover, \ourmethod{}'s robust performance across random and OOD settings demonstrates that its generated parameters effectively capture the stylistic and personal nuances crucial for generating high-quality and personalized responses.


\begin{figure}
    \centering
    \includegraphics[width=0.90\linewidth]{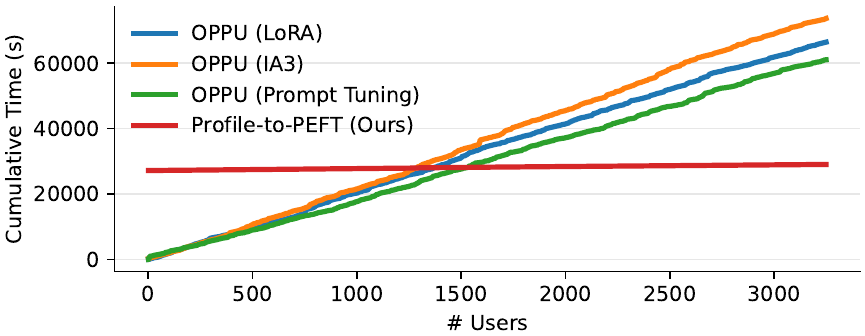}
    \caption{Cumulative time for personalized parameter generation vs. user count. Our Profile-to-PEFT exhibits near-constant, minimal cost, showing superior scalability over OPPU variants' linear growth.}
    \label{fig:efficiency}
    \vspace{-0.3cm}
\end{figure}

\section{Analysis}

\paragraph{Training User Quantity \emph{v.s.} Diversity}

We investigate how \ourmethod{}'s performance varies with training user quantity and diversity. Figure \ref{fig:scaling_users} displays performance on \textit{random} and \textit{OOD} splits, varying user diversity by controlling user clusters from 10 to 50 and user count from 20\% to 100\%.
Results indicate that increasing user quantity yields only marginal gains, with top-row performance curves remaining largely flat from 20\% to 100\% across all tasks. In contrast, increasing diversity positively impacts performance, as bottom-row curves show consistent improvements from 10 to 50 clusters across all task categories for both splits. For example, in classification tasks, OOD F1-score rises from approximately 0.508 to 0.560 with greater diversity, but shows no corresponding gain with increased quantity. These findings suggest that user profile diversity is more critical than sheer training user volume for robust, generalizable performance.

\paragraph{Deployment Efficiency}
We compare the time required to generate personalized PEFT parameters for each user at deployment. On average, OPPU (LoRA) takes 20.44 s per user, OPPU (IA3) takes 22.67 s per user, and OPPU (Prompt Tuning) takes 18.78 s per user. In contrast, our proposed \ourmethod{} requires only 0.57 s per user, representing a speedup of 33x compared to the fastest OPPU variant. Figure \ref{fig:efficiency} visually confirms this scalability: the cumulative time for OPPU methods increases steeply and linearly with the user count , while the cost for \ourmethod{} remains near-zero and constant. Although \ourmethod{} requires a one-time upfront training cost of 27167 seconds, this investment is amortized after ~1,450 users, making the framework substantially more efficient for large-scale, real-time applications.

\begin{table}[t]
   \caption{Performance of \ourmethod{} with different embedding models on classification, generation, and rating prediction tasks under OOD split.}
   \centering
   \small
   \begin{adjustbox}{max width=1\linewidth}
   \begin{tabular}{lcccccc}
   \toprule[1.2pt]
   \multirow{2}{*}{\textbf{Embedding Model}} & \multicolumn{2}{c}{\textbf{Class.} $\uparrow$} & \multicolumn{2}{c}{\textbf{Text Gen.} $\uparrow$} & \multicolumn{2}{c}{\textbf{Rating Pred.} $\downarrow$} \\
   \cmidrule(lr){2-3} \cmidrule(lr){4-5} \cmidrule(lr){6-7}
   & Acc & F1 & R-1 & R-L & MAE & RMSE \\
   \midrule[0.75pt]
   \texttt{Qwen3-Emb-0.6B} & .562 & .544 & \underline{.316} & \textbf{.244} & \textbf{.258} & \textbf{.543} \\
   \texttt{Qwen3-Emb-4B}& \textbf{.581} & \textbf{.562} & \textbf{.326} & \underline{.243} & \textbf{.258} & \underline{.583} \\
   \texttt{Qwen3-Emb-8B} & .560 & .544 & .313 & .234 & .391 & .715 \\
   \midrule
   \texttt{Qwen2.5-7B-It} & \underline{.571} & \underline{.552} & .311 & .241 & .281 & .596\\
   \texttt{gte-large-en} & .557 & .538 & .313 & .240 & \underline{.276} & .600 \\
   \midrule
   \textit{Non-Personalized} & .532 & .525 & .294 & .211 & .465 & .789 \\
   \bottomrule[1.2pt]
   \end{tabular}
   \end{adjustbox}
   \vspace{-0.3cm}
   \label{tab:embedding_models}
\end{table}

\paragraph{On Embedding Model Choice} 
The quality of user embeddings is critical for the hypernetwork. Our experiments show that all tested embedding backbones yield substantial improvements over the non-personalized baseline across all task categories. Among the dedicated embedding models, our default choice, \texttt{Qwen3-Emb-4B}, achieves the highest classification and text generation scores. Notably, the larger \texttt{Qwen3-Emb-8B} model underperforms its smaller counterparts across all metrics, suggesting that simply increasing embedding model size does not guarantee better performance. The framework also proves effective when using the base model's own last-layer activations (\texttt{Qwen2.5-7B-It}) as user embeddings. These results demonstrate that while the framework is robust to different embedding backbones, a high-quality, mid-sized model like \texttt{Qwen3-Emb-4B} provides the strongest overall performance.

\begin{figure}[t]
   \centering
   \includegraphics[width=1\linewidth]{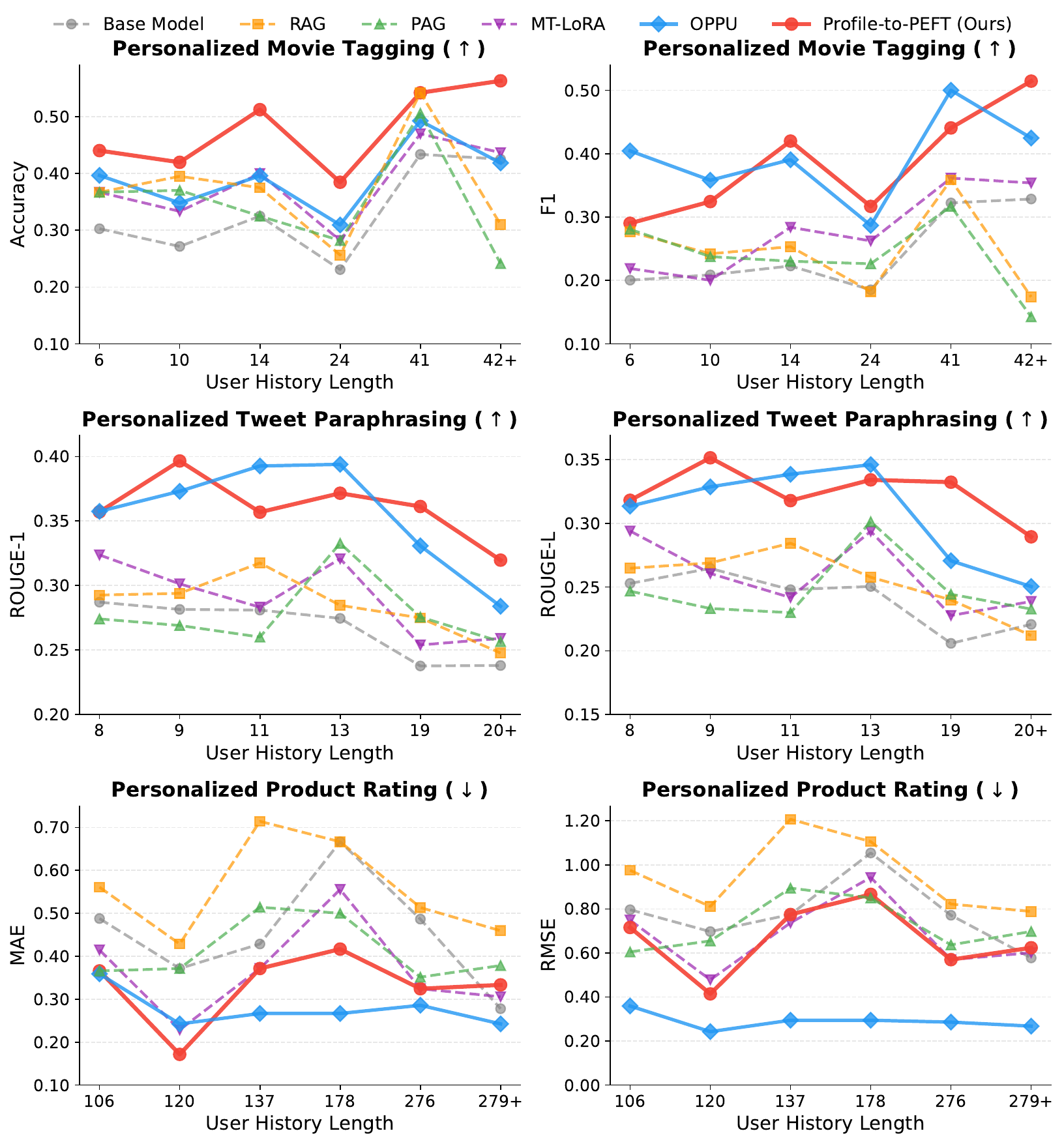}
   \caption{Robustness of \ourmethod{} to varying user activity levels. The model maintains strong performance across different history lengths, outperforming baselines for both sparse and dense user data.}
   \label{fig:active_levels}
   \vspace{-10pt}
\end{figure}

\paragraph{Performance \emph{w.r.t.} User Active Levels}
User engagement varies significantly. We analyze \ourmethod{}'s robustness across users with varying history lengths in three representative tasks under classification, generation, and rating prediction categories (Shown in Figure \ref{fig:active_levels}). The results demonstrate that \ourmethod{} consistently delivers strong performance regardless of the user's activity level. For all three categories, \ourmethod{} remains highly competitive with OPPU, which is fine-tuned directly on target user data, and outperforms other baselines across all history length buckets. Even for users with very sparse histories, \ourmethod{} effectively generates high-quality parameters, showcasing its robustness and suitability for real-world scenarios with diverse user engagement.

\paragraph{Ablation Study} We conducted systematic ablations to validate key design choices in \ourmethod{} and show results in Table \ref{tab:ablation}. Disrupting the personalization signal by providing a shuffled user profile causes a significant performance drop across all tasks, with the F1-score for classification falling from 0.562 to 0.521 and the MAE for rating prediction increasing from 0.258 to 0.322. This confirms that the hypernetwork effectively learns from the semantic content of the user profile rather than merely fitting to its structure. Further analysis of the user profile components reveals that the user summary is the most critical input. Using the summary alone achieves performance nearly on par with the full model (\emph{e.g.}, 0.562 \emph{vs.} 0.581 in classification accuracy). In contrast, relying solely on retrieved user history leads to a substantial decline in performance, particularly in rating prediction, where the MAE increases by over 56\% (from 0.258 to 0.405). These results underscore the importance of a high-quality user summary as the primary signal for generating effective personalized parameters.

\begin{table}[t]
  \vspace{-0.3cm}
  \caption{Component ablation study for our method under the \textit{OOD} split, validating their effectiveness.}
  \centering
  \small
  \begin{adjustbox}{max width=1\linewidth}
  \begin{tabular}{lcccccc}
  \toprule[1.2pt]
  \multirow{2}{*}{\textbf{Embedding Model}} & \multicolumn{2}{c}{\textbf{Class.} $\uparrow$} & \multicolumn{2}{c}{\textbf{Text Gen.} $\uparrow$} & \multicolumn{2}{c}{\textbf{Rating Pred.} $\downarrow$} \\
   \cmidrule(lr){2-3} \cmidrule(lr){4-5} \cmidrule(lr){6-7}
   & Acc & F1 & R-1 & R-L & MAE & RMSE \\
  \midrule[0.75pt]
  
  \textbf{\ourmethod{} (Full)} & \textbf{.581} & \textbf{.562} & \textbf{.326} & \textbf{.243} & \textbf{.258} & \textbf{.583} \\
  \midrule
  \midrule
  \multicolumn{7}{l}{\textbf{{\textit{Personalization Signal Ablations:}}}}\\
  \quad - random user profile &  \underline{.570} & \underline{.553} & .304 & .228 & \underline{.276} & .601 \\
  \quad - shuffle user profile & .535 & .521 & .307 & .223 & .322 & .692 \\
  \midrule
  \multicolumn{7}{l}{\textbf{{\textit{User Profile Contribution:}}}}\\
  \quad - user summary only &  .562 & .545 & \underline{.313} & \underline{.240} & .304 & \underline{.584}\\
  \quad - retrieved history only & .538 & .521 & .298 & .216 & .405 &.712 \\
  \quad - full history only & .541 & .526 & .302 & .217 & .392 & .740 \\
  
  \bottomrule[1.2pt]
  \end{tabular}
  \end{adjustbox}
  
  \vspace{-0.3cm}
  \label{tab:ablation}
\end{table}

\section{Related Work}
\paragraph{Personalization of LLMs}
Existing LLM personalization methods can be broadly categorized into prompt-based and Parameter-Efficient Fine-Tuning (PEFT)-based approaches. Prompt-based methods integrate user data into the model's input context. This includes using raw user history as few-shot examples \citep{dai2023uncovering, wang2023learning, kang2023llms,kim-yang-2025-shot}, retrieving relevant history snippets to overcome context limitations \citep{salemi2023lamp, mysore2023pearl, salemi2024optimization}, or augmenting queries with summarized user profiles \citep{richardson2023integrating, sun2024persona,dong-etal-2024-llm, tan2025perrecbench, zhang-2024-guided}. Further research has explored enhancing these methods with planning \citep{salemi2025lamp} and reasoning capabilities \citep{salemi2025reasoning}. In contrast, PEFT-based methods embed user preferences directly into lightweight model parameters. Notable examples include training one PEFT per user (OPPU) \citep{tan2024democratizing}, enabling collaborative personalization \citep{tan2024personalized}, or performing group-level adaptation \citep{zhang2025proper}. User-LLM \citep{ning2025user} learn user embeddings to contextualize LLMs for personalization. Another line of work focuses on personalized alignment through techniques like parameter merging \citep{jang2023personalized}, RLHF \citep{li2024personalized, park2024principled}, custom reward models \citep{cheng2023deserves,bose2025lore,shenfeld2025language}, and under black-box model \citep{zhuang2024hydra}, and conversational settings \citep{zhao2025personalens,wu-etal-2025-aligning}.


\paragraph{Hypernetwork for PEFT Generation} 
A hypernetwork, a neural network that generates weights for another model \citep{ha2016hypernetworks}, has become a key strategy for PEFT of LLMs. This approach produces task-specific modules without costly full-model retraining. Pioneering work like HyperFormer used a hypernetwork to generate adapter layers for different NLP tasks \citep{mahabadi2021parameter}. This concept has since been adapted for various LLM PEFT methods. For instance, some techniques generate soft prompts \citep{he2022hyperprompt}, while others create adapter weights from task embeddings \citep{phang2023hypertuning} or textual descriptions \citep{ivison2022hint}. More recent methods focus on generating LoRA parameters directly. While HyperLoRA conditions on few-shot examples \citep{lv2024hyperlora}, subsequent approaches like Text-to-LoRA \citep{charakorn2025text} and DnD \citep{liang2025drag} generate LoRA weights from natural language task descriptions or unlabeled prompts. This evolution enables efficient, on-the-fly adaptation and zero-shot generalization to new tasks without requiring explicit examples or task IDs.

While prior work uses hypernetworks for task-level adaptation, \ourmethod{} pioneers this approach for user-level personalization. It generates PEFT parameters directly from natural language user profiles or histories, enabling generalization to unseen users and real-time adaptation without per-user fine-tuning. This offers a practical and scalable solution for industrial PEFT-based LLM personalization.

\section{Conclusion}
We introduce \ourmethod{}, a hypernetwork-based framework that generates personalized PEFT parameters directly from user profiles. This approach produces customized LoRA adapters in a single inference pass at deployment, eliminating the costly per-user fine-tuning of traditional methods and enabling real-time updates. Experiments demonstrate that \ourmethod{} matches the performance of computationally intensive baselines and generalizes effectively and robustly to unseen users during deployment. By decoupling parameter generation from per-user training, \ourmethod{} offers a practical path toward deploying dynamic, privacy-preserving, and truly individualized LLMs at scale, paving the way for instantly adaptive personalized AI systems.

\section*{Limitations}
We identify two key limitations in \ourmethod{}. First, constrained by the dataset, our focus is primarily on one specific task per user rather than examining user behaviors across multiple tasks and domains. For instance, in the movie tagging task, users are solely engaged in that specific activity, without the inclusion of behaviors from other domains or platforms. Despite this, the \ourmethod{} framework is inherently adaptable to any text sequence generation task and is compatible with diverse user instructions across various tasks and domains. Personalizing LLM across a broader range of tasks and domains is left as future work.
Second, despite our proposed \ourmethod{} is compatible with all PEFT methods that introduce trainable modules throughout the model, such as Adapter \citep{houlsby2019parameter}, $\mathrm{(IA)^3}$ \citep{liu2022few}, and prefix tuning \citep{li-liang-2021-prefix}, we primarily focus on LoRA in this work. This is due to LoRA's popularity, widespread use, and superior performance demonstrated by OPPU \citep{tan2024democratizing}, while we expect to expand our experiment and analysis to more PEFT methods in future work.

\section*{Ethical Considerations}
\paragraph{Privacy}
The Profile-to-PEFT framework is designed to enhance user privacy by enabling local deployment, where personalized parameters are generated without transmitting raw user history to a central server. This significantly mitigates the risk of data leakage during transmission. However, the generated PEFT parameters themselves are a compressed representation of a user's profile and preferences. This raises a potential concern that the parameters could be reverse-engineered to infer sensitive information about the user. Therefore, ensuring the security and privacy of these generated adapter weights is crucial, especially if they are stored or managed by a service provider.

\paragraph{Data Bias and Manipulation}
The personalization process inherently relies on a user's historical data. If this data contains existing biases, stereotypes, or prejudices, the hypernetwork will learn to encode these undesirable patterns into the personalized PEFT parameters. This could lead to the LLM generating outputs that reinforce or amplify a user's biases, creating a harmful echo chamber. Furthermore, the ability to directly map a profile to model behavior could be exploited for malicious manipulation, where a crafted profile could generate parameters that cause the LLM to subtly persuade or mislead a user. It is essential to develop methods for auditing and mitigating bias in both the input user data and the resulting personalized models.

\paragraph{Accessibility and Responsibility}
While Profile-to-PEFT significantly lowers the computational barrier for deploying personalized models compared to the "One-PEFT-Per-User" approach, the initial training of the hypernetwork on a diverse user population remains a resource-intensive task. This could still present an accessibility challenge for smaller organizations or researchers, potentially concentrating the power to create such systems in the hands of a few large entities. Developers of this technology have a responsibility to consider these downstream effects and to implement safeguards that prevent the system from being used for harmful purposes, such as generating discriminatory content or facilitating large-scale manipulation.

\section*{Acknowledgements}
This work was partially supported by NSF IIS-2119531, IIS-2137396, IIS-2142827, IIS-2234058, and Coefficient Giving. We also appreciate the support from the Foundation Models and Applications Lab of Lucy Institute and ND-IBM Tech Ethics Lab.


\newpage
\bibliography{custom}

\newpage
\clearpage
\appendix

\begin{figure*}[t]
   \centering
   \includegraphics[width=1\linewidth]{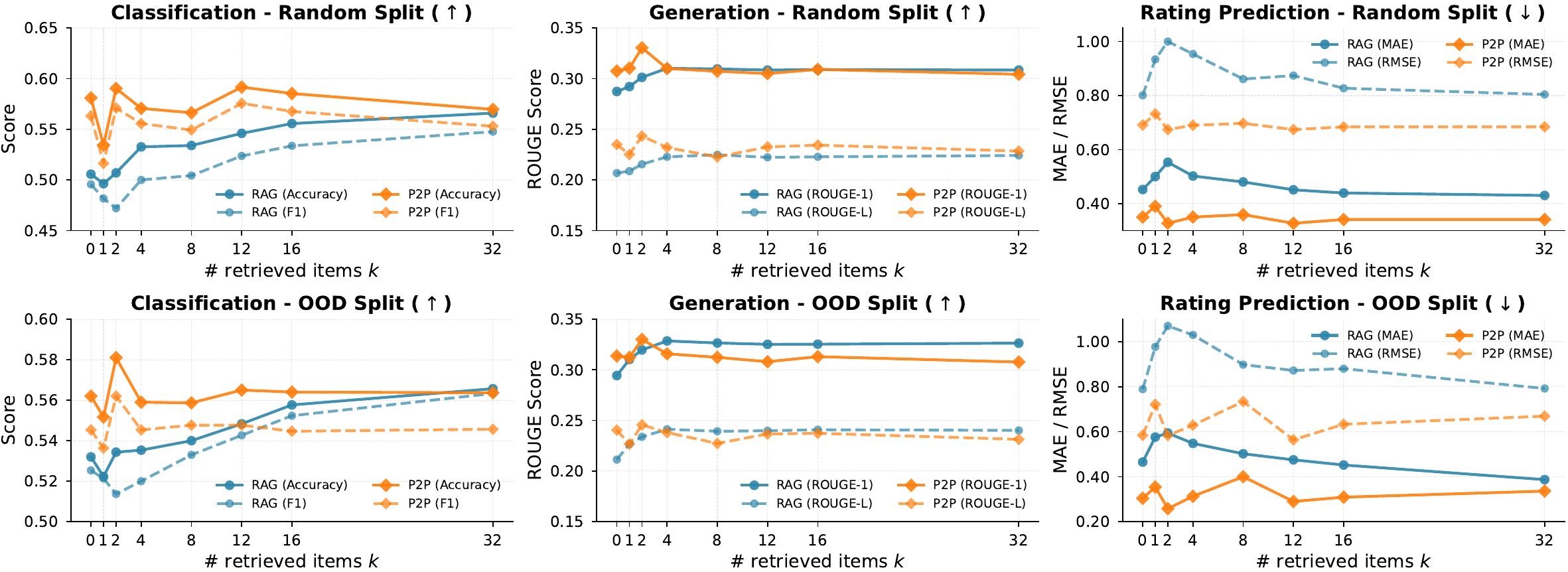}
   \caption{Performance of \ourmethod{} and RAG baseline method with different retrieval item $k$ under both \textit{Random} and \textit{OOD} split.}
   \label{fig:retrieval_k}
\end{figure*}

\section{Performance \emph{w.r.t.} Retrieval Top $k$}

We analyze the impact of the number of retrieved historical items $k$ on the performance of both \ourmethod{} and the RAG baseline across classification, generation, and rating prediction tasks. As shown in the Figure \ref{fig:retrieval_k}, the performance of \ourmethod{} remains remarkably stable and consistently high across all values of $k$, from 0 to 32. This holds true for both the random and OOD splits. The flat performance curves indicate that our method is not sensitive to the number of retrieved items, suggesting that the user summary provides a strong, condensed personalization signal that makes the model robust and less reliant on a dynamic retrieval step for each input. In contrast, the performance of the RAG baseline is highly dependent on $k$. For classification and generation tasks, RAG's performance generally improves as more items are retrieved, though it often plateaus or slightly degrades with a very high number of items. This highlights RAG's sensitivity to the quality and quantity of the retrieved context. Across all tasks and splits, \ourmethod{} consistently outperforms the RAG baseline, demonstrating the superiority of encoding user preferences into the model's parameters over simply augmenting the input prompt.

\begin{table*}[t]
    \caption{Main experiment results on the LaMP and LongLaMP benchmarks under the \textit{random split} setting using \texttt{Qwen2.5-3B-it} as base model. $\uparrow$ indicates that higher values are better, and $\downarrow$ implies lower values are preferable. For each task, the best score is in \textbf{bold} and the second best is \underline{underlined}. The final row reports average per-instance inference time (ms).}
    \centering
    \begin{adjustbox}{max width=0.9\linewidth}
        \begin{tabular}{llcccc|cc|c}
        \toprule[1.5pt]
        \multirow{2}{*}{\textbf{Task}} & \multirow{2}{*}{\textbf{Metric}} & \multirow{2}{*}{\makecell{\textbf{Base}\\\textbf{Model}}} & \multirow{2}{*}{\textbf{RAG}} & \multirow{2}{*}{\textbf{PAG}} & \multirow{2}{*}{\makecell{\textbf{Full}\\\textbf{History}}} & \multirow{2}{*}{\makecell{\textbf{MT}\\\textbf{LoRA}}} & \multirow{2}{*}{\textbf{OPPU}} & \multirow{2}{*}{\makecell{\textbf{\ourmethod{}}\\ \textbf{(Ours)}}} \\
        & & & & & & & \\
        \midrule[0.75pt]

        \multirow{2}{*}{\makecell[l]{\textsc{LaMP-1:}\\\textsc{Citation Id.}}} & Acc $\uparrow$ & .456	& .543 & .574 & \underline{.582} & .511	& .480 & \textbf{.591} \\
        & F1 $\uparrow$ & .421	& .530	&.574	& \underline{.577}	&.490	&.475	& \textbf{.585} \\
        \hline
        \multirow{2}{*}{\makecell[l]{\textsc{LaMP-2N: }\\\textsc{News Cat.}}} & Acc $\uparrow$& .544	& .630	&\textbf{.724}	&\underline{.701}	&.655	&.605	&.69 \\
        & F1 $\uparrow$& .575	&.652	&\textbf{.741}	&\underline{.719}	&.644	&.635 & .683\\
        \hline
        \multirow{2}{*}{\makecell[l]{\textsc{LaMP-2M: }\\\textsc{Movie Tagging}}} & Acc $\uparrow$&.247	&.326	&.332	&\underline{.384}	&.307	&.226	&\textbf{.413} \\
        & F1 $\uparrow$& .262	&.349	&.344	&\underline{.397}	&.298	&.244	&\textbf{.402}\\
        \hline
        \multirow{2}{*}{\makecell[l]{\textsc{LaMP-3: }\\\textsc{Product Rating}}} & MAE $\downarrow$& .941	&.859	&.665	&\underline{.475}	&.495	&.656	&\textbf{.405} \\
        & RMSE $\downarrow$&1.31&	1.27	& 1.08	& .889	&\underline{.882}	&1.04	&\textbf{.794}\\
        \hline
        \multirow{2}{*}{\makecell[l]{\textsc{LaMP-4:}\\\textsc{News Headline Gen.}}} & R-1 $\uparrow$&.119	&.137	&.129	&\underline{.147}	&.141	&.132	&\textbf{.155}\\
        & R-L $\uparrow$& .109	&.125	&.117	& \underline{.132}	&.128	&.118	& \textbf{.141}\\
        \hline
        \multirow{2}{*}{\makecell[l]{\textsc{LaMP-5: }\\\textsc{Scholarly Title Gen.}}} & R-1 $\uparrow$&  .443	&\underline{.474}	&.423	&.468	&.454	&.449	&\textbf{.484}\\
        & R-L $\uparrow$& .379	&\underline{.412}	&.362	&.407	&\underline{.412}	&.384	&\textbf{.438} \\
        \hline
        \multirow{2}{*}{\makecell[l]{\textsc{LaMP-7:}\\\textsc{Tweet Paraphrase}}} & R-1 $\uparrow$& .370 &	.377	&.364	&.383 & \textbf{.466}	&.321	& \underline{.462}\\
        & R-L $\uparrow$& .316	&.322	&.308	&.323	&\textbf{.403}	&.270 & \underline{.397}\\
        \hline
        \multirow{2}{*}{\makecell[l]{\textsc{LongLaMP-1:}\\\textsc{Abstract Gen.}}} & R-1 $\uparrow$& .312 & \textbf{.348}	&.337	&\underline{.347}	&.343	&.304	&.346\\
        & R-L $\uparrow$& .170	&.189	&.187	&\underline{.191}	&.190	&.161	&\textbf{.200}\\
        \hline
        \multirow{2}{*}{\makecell[l]{\textsc{LongLaMP-2:}\\\textsc{Topic Writing}}} & R-1 $\uparrow$& .226	& \underline{.249}	&\textbf{.271}	&.246	&.251	&.237	&.204\\
        & R-L $\uparrow$& .115	& .127	& \textbf{.134}	& \underline{.129}	& .128	& .114	&.117 \\
        \hline
        \multirow{2}{*}{\makecell[l]{\textsc{LongLaMP-3:}\\\textsc{Review Writing}}} & R-1 $\uparrow$& .191	& \underline{.233}	&\textbf{.271}	&.223	&.229	&.202	&.222\\
        & R-L $\uparrow$& .112	&.126	&\textbf{.137}	&.124	&.129	&.113	&\underline{.131}\\
        \midrule[0.75pt]
        \multicolumn{9}{c}{\textit{Average Performance}} \\
        \midrule[0.75pt]
        \multirow{2}{*}{\makecell[l]{\textsc{Classification} }} & Acc $\uparrow$ & .415 & .499 & .543 & \underline{.556} & .491 & .437 & \textbf{.565}\\
        & F1 $\uparrow$ & .419 & .510 & .553 & \textbf{.564} & .477 & .451 & \underline{.557}\\
        \hline
        \multirow{2}{*}{\makecell[l]{\textsc{Generation}}} & R-1 $\uparrow$ & .277 & .303 & .299 & .302 & \textbf{.314} & .274 & \underline{.312}\\
        & R-L $\uparrow$ & .200 & .217 & .208  & .218 & \underline{.232} & .193 & \textbf{.237} \\
        \hline
        \hline
        \textsc{Infer. Time} & ms $\downarrow$& 21.94 & 31.58 & 43.25 & 258.66 & 20.86 & 24.13 & 27.51\\
        \bottomrule[1.5pt]
        \end{tabular}%
        \end{adjustbox}
     
        \label{tab:random_results_3b}
\end{table*}

\begin{table*}[t]
    \caption{Main experiment results on the LaMP and LongLaMP benchmarks under the \textit{OOD split} setting using \texttt{Qwen2.5-3B-it} as base model. $\uparrow$ indicates that higher values are better, and $\downarrow$ implies lower values are preferable. For each task, the best score is in \textbf{bold} and the second best is \underline{underlined}. The final row reports average per-instance inference time (ms).}
    \centering
    \begin{adjustbox}{max width=0.9\linewidth}
        \begin{tabular}{llcccc|cc|c}
        \toprule[1.5pt]
        \multirow{2}{*}{\textbf{Task}} & \multirow{2}{*}{\textbf{Metric}} & \multirow{2}{*}{\makecell{\textbf{Non-}\\\textbf{Personalized}}} & \multirow{2}{*}{\textbf{RAG}} & \multirow{2}{*}{\textbf{PAG}} & \multirow{2}{*}{\makecell{\textbf{Full}\\\textbf{History}}} & \multirow{2}{*}{\makecell{\textbf{MT}\\\textbf{LoRA}}} & \multirow{2}{*}{\textbf{OPPU}} & \multirow{2}{*}{\makecell{\textbf{\ourmethod{}}\\ \textbf{(Ours)}}} \\
        & & & & & & & \\
        \midrule[0.75pt]

        \multirow{2}{*}{\makecell[l]{\textsc{LaMP-1: Personalized}\\\textsc{Citation Identification}}} & Acc $\uparrow$ & .491 & .560 &	\underline{.580}	&\textbf{.588}	&.505	&.502	&.573\\
        & F1 $\uparrow$ & .455	&.555	&\textbf{.580}	&.586	&.491	&.502	& \underline{.570} \\
        \hline
        \multirow{2}{*}{\makecell[l]{\textsc{LaMP-2N: Personalized}\\\textsc{News Categorization}}} & Acc $\uparrow$& .437	&.550	&.544	&\underline{.583}	&.519	&.495	&\textbf{.596} \\
        & F1 $\uparrow$& .471	& \underline{.582}	&.567	&\textbf{.591}	&.505	&.522	&\underline{.582}\\
        \hline
        \multirow{2}{*}{\makecell[l]{\textsc{LaMP-2M: Personalized}\\\textsc{Movie Tagging}}} & Acc $\uparrow$& .271	&.376	&.374	&\textbf{.476}	&.327 &	.203	&\underline{.430}\\
        & F1 $\uparrow$& .308	&.419	&.419	&\textbf{.507}	&.338&	.239	&\underline{.432}\\
        \hline
        \multirow{2}{*}{\makecell[l]{\textsc{LaMP-3: Personalized}\\\textsc{Product Rating}}} & MAE $\downarrow$& 1.00	&.802	&.571	&\underline{.327}	&.387	&.635	&\textbf{.239} \\
        & RMSE $\downarrow$&1.40 &	.126	&.979	&\underline{.715}	&.803	&.107	&\textbf{.599}\\
        \hline
        \multirow{2}{*}{\makecell[l]{\textsc{LaMP-4: Personalized}\\\textsc{News Headline Gen.}}} & R-1 $\uparrow$& .151	&.184	&.172	&\underline{.188}	&.179	&.151	&\textbf{.193}\\
        & R-L $\uparrow$& .133	&.164	&.153	&\underline{.168}	&.158	&.134	&\textbf{.175}\\
        \hline
        \multirow{2}{*}{\makecell[l]{\textsc{LaMP-5: Personalized}\\\textsc{Scholarly Title Gen.}}} & R-1 $\uparrow$&  .450	& .467	&.434	&\textbf{.477}	&.461	&.454	&\underline{.476}\\
        & R-L $\uparrow$& .395	&.407	&.383	&\underline{.421}	&.420	&.397	&\textbf{.432}\\
        \hline
        \multirow{2}{*}{\makecell[l]{\textsc{LaMP-7: Personalized}\\\textsc{Tweet Paraphrasing}}} & R-1 $\uparrow$& .384	&.441	&.447	&.449	&\textbf{.483}	&.358	&\underline{.453}\\
        & R-L $\uparrow$& .320	&.384	&.389	&\underline{.395}	&\textbf{.406}	&.299	&.382\\
        \hline
        \multirow{2}{*}{\makecell[l]{\textsc{LongLaMP-1:}\\\textsc{Abstract Generation}}} & R-1 $\uparrow$&.306	& \underline{.340}	&.336	&\textbf{.342}	&.331	&.295	&.330\\
        & R-L $\uparrow$& .164	& \underline{.188}	&.185	&.184	&.183	&.156	&\textbf{.189}\\
        \hline
        \multirow{2}{*}{\makecell[l]{\textsc{LongLaMP-2:}\\\textsc{Topic Writing}}} & R-1 $\uparrow$& .228	& \underline{.251}	& \textbf{.274} &.240	&.225	&.232	&.206\\
        & R-L $\uparrow$& .115	&\underline{.135}	&\textbf{.140}	&.132	&.120	&.111	&.119\\
        \hline
        \multirow{2}{*}{\makecell[l]{\textsc{LongLaMP-3:}\\\textsc{Review Writing}}} & R-1 $\uparrow$&.175	& \underline{.223}	&\textbf{.261}	&.213	&.209	&.195	&.204  \\
        & R-L $\uparrow$& .102	&\underline{.121}	&\textbf{.133}	&.119	&.118	&.107	&.120 \\
        \midrule[0.75pt]
        \multicolumn{9}{c}{\textit{Average Performance}} \\
        \midrule[0.75pt]
        \multirow{2}{*}{\makecell[l]{\textsc{Classification} }} & Acc $\uparrow$ &  .400 & .495 & .499 & \textbf{.549} & .450 & .400 & \underline{.533}\\
        & F1 $\uparrow$ & .411 & .518 & \underline{.522} &.451 & .445 & .421 & \textbf{.528}\\
        \hline
        \multirow{2}{*}{\makecell[l]{\textsc{Generation}}} & R-1 $\uparrow$ & .282 & \underline{.318} & \textbf{.321} & .318 & .315 & .280 & .310 \\
        & R-L $\uparrow$ & .205 & .233 & .231 & \textbf{.236} & \underline{.234} & .201 & \textbf{.236} \\
        \hline
        \hline
        \textsc{Infer. Time} & ms $\downarrow$& 22.17 & 50.59 & 33.76 & 223.24 & 22.43 & 24.91 & 26.83 \\
        \bottomrule[1.5pt]
        \end{tabular}%
        \end{adjustbox}
     
        \label{tab:ood_results_3b}
\end{table*}

\section{Performance with Additional Base Model}
To evaluate the robustness of our framework, we replicated our experiments using a smaller base model, \texttt{Qwen2.5-3B-it}. The results, presented in Table \ref{tab:random_results_3b} and \ref{tab:ood_results_3b}, demonstrate that the core advantages of \ourmethod{} are consistent across different model sizes.

Even with the smaller 3B model, \ourmethod{} consistently outperforms the computationally expensive OPPU baseline and remains highly competitive with strong prompt-based methods in both the \textit{random} and \textit{OOD} splits. In the random split, our method achieves the highest average accuracy (0.565) and F1-score (0.557) in classification tasks. Similarly, in the more challenging OOD split, it maintains a significant lead over other parameter-based methods like OPPU and MT-LoRA, showcasing its strong generalization capabilities.

While the absolute performance scores are naturally slightly lower than those achieved with the larger 7B model, the relative performance gains and overall trends remain the same. This confirms that the effectiveness of the \ourmethod{} framework is not contingent on a large-scale base model and that our approach provides a robust and scalable solution for LLM personalization.

\section{Baseline Details}
\label{app:baseline}
We present the details of baseline methods to help facilitate reproducibility.
\begin{itemize}[leftmargin=*]
    \item \textbf{Retrieval-Augmented Generation (RAG)} \cite{salemi2023lamp}: The LLM input is appended with the top $k$ relevant user history items \emph{w.r.t.} the user input $x_u$, where the input sequence $x'$ can be defined as 
    \begin{align*}
        x_u'=[\mathcal{R}(x_u, \mathcal{H}_u, k) \; || \; x_u],
    \end{align*}
    where $\mathcal{R}$ is the retriever, default to BM25. $[\cdot||\cdot]$ denotes the concatenation operation. 
    \item \textbf{Profile-Augmented Generation (PAG)} \cite{richardson2023integrating}: The LLM input is appended with user summary $s_u$ and top $k$ retrieved user history items that most relevant to user input $x_u$. The user summary $s_u$ is generated by the base model from the sampled user history entries. The input of LLM $x'$ is represented as
    \begin{align*}
        x_u'=[s_u \; || \; \mathcal{R}(x_u, \mathcal{H}_u, k) \; || \; x_u].
    \end{align*}
    \item \textbf{Full History}: in this method, the user context is the entire user history corpus. Given the long context window of 32k tokens, the model can consume almost all of the historical data. If the flattened user history overflow the context length, we keep the most recent history items that fit in the context length. The input of LLM $x'$ is represented as
    \begin{align*}
        x_u'=[\mathrm{Flatten}(\mathcal{H}_u)\; || \; x_u].
    \end{align*}
    
    \item \textbf{Multi-task LoRA (MT-LoRA)}: The multi-task LoRA method serves as a task-level adaptation without user context for personalization. It trains a LoRA using the input and output pairs from the training set without including user profile or user history for personalization signals. The training objective can be represented as:
    \begin{align*}
         \Delta W^* = \arg\min_{\Delta W} \mathcal{L}_{\text{SFT}}(\Psi \oplus \Delta W, \{(x_u, y_u)\}),
    \end{align*}
   where the $\Psi$ is the base model parameters, $\{(x_u, y_u)\}$ denotes the user input and output pair in training set for SFT training.
   
    \item \textbf{One PEFT Per User (OPPU)} \cite{tan2024democratizing}: OPPU trains the per-user PEFT parameters on the target user history from scratch. The personalized PEFT parameters for user $u$ can be represented as:
    \begin{align*}
        \Delta W_u^* = \arg\min_{\Delta W} \mathcal{L}_{\text{SFT}}(\Psi \oplus \Delta W, \mathcal{H}_u^{<t}),
    \end{align*}
    The $\Delta W_u$ requires per-user training using the target user history data from scratch and needs iteratively went through the entire user history corpus using backpropagation to optimize the PEFT parameters, making it request extensive computation as the number of users scale up.

\end{itemize}

\section{Task Details}
\label{app:tasks}
We present the task details as follows to help readers gain a better understanding of the task format.
\begin{itemize}[leftmargin=*]
    \item \textbf{LaMP-1: Personalized Citation Identification} is a binary text classification task. Specifically, given user $u$ writes a paper $x$, the task aims to make the model determine which of the two candidate papers $u$ will cite in paper $x$ based on the user's history data, which contains the publications of user $u$.

    \item \textbf{LaMP-2N: Personalized News Categorization} is a 15-way text classification task to classify news articles written by a user $u$. Formally, given a news article $x$ written by user $u$, the language model is required to predict its category from the set of categories based on the user's history data, which contains the user's past article and corresponding category.

    \item \textbf{LaMP-2M: Personalized Movie Tagging} is a 15-way text classification task to make tag assignments aligned with the user's history tagging preference. Specifically, given a movie description $x$, the model needs to predict one of the tags for the movie $x$ based on the user's historical movie-tag pairs.

    \item \textbf{LaMP-3: Personalized Product Rating} is a 5-way text classification task and can also be understood as a regression task. Given the user $u$'s historical review and rating pairs and the input review $x$, the model needs to predict the rating corresponding to $x$ selected from 1 to 5 in integer. 

    \item \textbf{LaMP-4: Personalized News Headline Generation} is a text generation task to test the model's ability to capture the stylistic patterns in personal data. Given a query $x$ that requests to generate a news headline for an article, as well as the user profile that contains the author's historical article-title pairs, the model is required to generate a news headline specifically for the given user.

    \item \textbf{LaMP-5: Personalized Scholarly Title Generation} is a text generation task to test personalized text generation tasks in different domains. In this task, we require language models to generate titles for an input article $x$, given a user profile of historical article-title pairs for an author.

    \item \textbf{LaMP-7: Personalized Tweet Paraphrasing} is also a text generation task that tests the model's capabilities in capturing the stylistic patterns of authors. Given a user input text $x$ and the user profile of historical tweets, the model is required to paraphrase $x$ into $y$ that follows the given user's tweet pattern.
    \item \textbf{LongLaMP-1: Abstract Generation} is a long-form text generation task designed to test personalized summarization. Given a user's history of writing academic papers and the body of a new, unseen paper, the model is required to generate an abstract for the new paper that aligns with the user's personal writing style.
    \item \textbf{LongLaMP-2: Topic Writing} is a long-form text generation task that assesses the model's ability to adopt a user's unique writing voice. Given a specific topic and a user profile containing their past writings, the model must generate a new piece of text on the given topic that emulates the user's personal style.
    \item \textbf{LongLaMP-3: Review Writing} is a long-form generation task focused on capturing a user's personal voice and opinion patterns. Given a product or business and a user's history of past reviews, the model is tasked with generating a new review that reflects the user's characteristic style and rating tendencies.
     \item \textbf{Empathetic Conversation} \citep{omitaomu2022empathic}: consists of 1000 essay responses (both empathy score and textual response) to a news article with their demographics and self-reported personality traits. It further includes dialog interactions between paired participants, enriched with various dialog annotations, such as other-reported empathy levels and turn-by-turn emotion ratings. This dataset can be used as both multiple-choice question answering and text generation dataset.
    \item \textbf{Personal Reddit} \citep{staab2023beyond}: consists of 500 samples of Reddit posts with their (anonymized) personal attributes, such as location, income, and sex. It contains user profile, question, and a ground-truth response given by the user, which can be used in our SFT training.
\end{itemize}

\section{Experimental Details}
To promote task diversity during \ourmethod{} training, each batch contains four different personalization tasks, with sampling weights proportional to the square root of each task’s dataset size to limit oversampling of small datasets. We train \ourmethod{} with a learning rate of $2\times10^{-5}$ for 20,000 steps and a batch size of 32 by default. For the generated LoRA adapters, we set the rank $r$=8 and insert trainable parameters into \texttt{q\_proj} and \texttt{v\_proj}. For inference, we use greedy decoding with temperature $\tau=0$ to reduce randomness and improve reproducibility.

\begin{table*}[t!]
\centering
\caption{Detailed statistics for all dataset splits. For each file, we report the number of users, number of queries, and the average length of input contexts and output generations (in character).}
\label{tab:dataset_stats_detailed}
\resizebox{0.8\textwidth}{!}{%
\begin{tabular}{@{}lllrrrr@{}}
\toprule
\textbf{Dataset} & \textbf{Subtask} & \textbf{Split} & \textbf{Users} & \textbf{Queries} & \textbf{Avg. Input Len} & \textbf{Avg. Output Len} \\
\midrule
\multirow{21}{*}{\textsc{LaMP}} & \multirow{3}{*}{\textsc{Citation Id.}} & train & 5947 & 7289 & 355.25 & 3.00 \\
 &  & ood\_test & 200 & 255 & 377.62 & 3.00 \\
 &  & random\_test & 200 & 254 & 358.93 & 3.00 \\
\cmidrule(l){2-7}
 & \multirow{3}{*}{\textsc{Movie Tagging}} & train & 733 & 5511 & 602.91 & 9.76 \\
 &  & ood\_test & 93 & 409 & 602.64 & 7.62 \\
 &  & random\_test & 92 & 518 & 602.92 & 9.36 \\
\cmidrule(l){2-7}
 & \multirow{3}{*}{\textsc{News Cat.}} & train & 253 & 7721 & 457.15 & 9.47 \\
 &  & ood\_test & 32 & 1106 & 445.29 & 8.90 \\
 &  & random\_test & 32 & 823 & 449.61 & 9.53 \\
\cmidrule(l){2-7}
 & \multirow{3}{*}{\textsc{News Headline Gen.}} & train & 1396 & 12147 & 178.02 & 62.50 \\
 &  & ood\_test & 100 & 487 & 219.65 & 49.77 \\
 &  & random\_test & 100 & 1125 & 174.10 & 63.62 \\
\cmidrule(l){2-7}
 & \multirow{3}{*}{\textsc{Product Rating}} & train & 19244 & 21658 & 694.97 & 1.00 \\
 &  & ood\_test & 200 & 217 & 493.73 & 1.00 \\
 &  & random\_test & 200 & 221 & 739.47 & 1.00 \\
\cmidrule(l){2-7}
 & \multirow{3}{*}{\textsc{Scholarly Title Gen.}} & train & 14274 & 15738 & 1094.14 & 76.60 \\
 &  & ood\_test & 200 & 218 & 1156.29 & 74.27 \\
 &  & random\_test & 200 & 217 & 1065.66 & 75.35 \\
\cmidrule(l){2-7}
 & \multirow{3}{*}{\textsc{Tweet Paraphrase}} & train & 12912 & 14346 & 180.44 & 93.16 \\
 &  & ood\_test & 200 & 225 & 184.37 & 93.75 \\
 &  & random\_test & 200 & 223 & 178.26 & 92.32 \\
\midrule
\multirow{9}{*}{\textsc{LongLaMP}} & \multirow{3}{*}{\textsc{Abstract Generation}} & train & 22103 & 30950 & 261.17 & 1116.49 \\
 &  & ood\_test & 200 & 283 & 256.51 & 1098.07 \\
 &  & random\_test & 200 & 277 & 259.04 & 1106.19 \\
\cmidrule(l){2-7}
 & \multirow{3}{*}{\textsc{Product Review}} & train & 15490 & 18928 & 725.36 & 1652.50 \\
 &  & ood\_test & 200 & 256 & 745.18 & 1813.55 \\
 &  & random\_test & 200 & 251 & 740.06 & 1412.50 \\
\cmidrule(l){2-7}
 & \multirow{3}{*}{\textsc{Topic Writing}} & train & 15330 & 19897 & 196.90 & 1438.58 \\
 &  & ood\_test & 200 & 279 & 184.61 & 1410.62 \\
 &  & random\_test & 200 & 273 & 208.48 & 1381.02 \\
\midrule
\multirow{3}{*}{\textsc{Personal Reddit}} & \multirow{3}{*}{-} & train & 409 & 409 & 394.66 & 656.11 \\
 &  & ood\_test & 53 & 53 & 385.45 & 602.70 \\
 &  & random\_test & 52 & 52 & 376.73 & 590.27 \\
\midrule
\multirow{3}{*}{\textsc{Empathetic Conv.}} & \multirow{3}{*}{-} & train & 53 & 712 & 4555.78 & 402.21 \\
 &  & ood\_test & 10 & 153 & 4307.86 & 487.34 \\
 &  & random\_test & 10 & 109 & 4438.62 & 385.30 \\
\bottomrule
\end{tabular}%
}
\label{tab:statistics}
\end{table*}

\section{Computational Resources}
\label{app:computation}
All the training experiments in this paper were conducted on a single node with 8 × NVIDIA A100-SXM4-80GB GPUs and Intel(R) Xeon(R) Platinum 8275CL CPU @ 3.00GHz.

\section{Scientific Artifacts}
\ourmethod{} is built with the help of many existing scientific artifacts, including PyTorch \citep{paszke2019pytorch}, Numpy \citep{harris2020array},  rank-bm25 \citep{rank_bm25}, and huggingface transformers \citep{wolf2020transformers}. We use vllm \citep{kwon2023efficient} as the inference framework. In our experiments, we use base model from \texttt{Qwen2.5} series and embedding models from \texttt{Qwen3-Emb} series, all models we used are released under apache-2.0 license. We will make the \ourmethod{} implementation publicly available to facilitate further research.

\section{Dataset Splits Details}
\label{app:split}
To ensure a representative random test split, we adopt a clustering-based diverse user selection strategy that leverages user embeddings to capture underlying population structures. Specifically, K-means clustering is applied to the normalized embeddings, with the number of clusters adaptively set between a minimum and maximum based on dataset size to achieve balanced group sizes. Cluster proportions are computed, and users are sampled proportionally from each cluster using random selection within clusters, ensuring the selected subset mirrors the overall distribution while incorporating randomness for variability; adjustments are made for rounding errors and small clusters to meet the exact target user count. In contrast, for the OOD test split, an extreme clustering approach is employed to maximize dissimilarity from the training population. The optimal number of clusters is determined via silhouette score maximization over a searched range, after which clusters are analyzed for size, intra-cluster compactness (average distance to centroid), and inter-cluster isolation (average distance to other centroids). A scoring metric prioritizing small, isolated clusters (defined as the inter-cluster distance divided by (1 + cluster size) guides the selection: entire high-scoring (outlier) clusters are allocated to the OOD test if they fit within the target size, or the farthest users from the centroid are chosen otherwise, enforcing strict cluster separation such that no training users originate from OOD-assigned clusters, thereby enhancing the OOD set's distinctiveness.
\section{Dataset Statics}
\label{app:statics}
The dataset statistics are presented in Table \ref{tab:statistics}.


\section{Prompt Details}
We present the prompt template for user profile generation and LLM-as-a-Judge evaluation.
\newpage

\begin{prompt}{Reward Scoring Prompt Template}
\#\#\#Task Description: \\
An instruction (might include an Input inside it), a response to evaluate, a reference answer that can get a score of 5, a user profile containing user preferences and information, and a score rubric representing a evaluation criteria are given. \\
1. Write a detailed feedback that assess the quality of the response strictly based on the given score rubric, not evaluating in general.\\
2. Consider how well the response aligns with the user's preferences, interests, and background information provided in the user profile when evaluating personalization quality.\\
3. After writing a feedback, write a score that is an integer between 1 and 5. You should refer to the score rubric.\\
4. The output format should look as follows: "(write a feedback for criteria) [RESULT] (an integer number between 1 and 5)"\\
5. Please do not generate any other opening, closing, and explanations.\\

\#\#\#The instruction to evaluate:\\
\{\{ instruction \}\}\\

\#\#\#Response to evaluate:\\
\{\{ response \}\}\\

\#\#\#Reference Answer (Score 5):\\
\{\{ reference\_answer \}\}\\

\#\#\#User Profile:\\
\{\{ user\_profile \}\}\\

\#\#\#Score Rubrics:\\
\{\{ rubric \}\}\\

\#\#\#Feedback:
\end{prompt}

\begin{prompt}{Rubric Template}
[\texttt{\{criteria\}}]\\
Score 1: \texttt{\{score1\_description\}}\\
Score 2: \texttt{\{score2\_description\}}\\
Score 3: \texttt{\{score3\_description\}}\\
Score 4: \texttt{\{score4\_description\}}\\
Score 5: \texttt{\{score5\_description\}}
\end{prompt}

\begin{prompt}{Score Rubric}
\texttt{criteria}:"Evaluate how well the response to the instruction is personalized to the specific user."\\
  \texttt{score1\_description}: "Generic or impersonal. Ignores the provided profile/personality. Style does not match the user; may feel robotic or off-topic. Makes incorrect assumptions or contradicts stated preferences. No meaningful use of user details; largely boilerplate."\\
  \texttt{score2\_description}: "Minimal personalization. Mentions a profile detail superficially but remains mostly generic. Weak style match; limited relevance to the user's interests or situation. Includes filler or distracting disclaimers. Significant deviation from the reference's intent or emphasis."\\
  \texttt{score3\_description}: "Basic personalization. References a few relevant details and partially adapts tone. Generally on topic but misses important user nuances (interests, constraints, or personality cues). Moderate similarity to the reference; may be verbose or somewhat generic." \\
  \texttt{score4\_description}: "Good personalization. Integrates multiple user details accurately; content is relevant and helpful. Tone largely matches the user's personality and preferred style. Clear, concise, and engaging with only minor misses versus the user's preferences or the reference's intent."\\
  \texttt{score5\_description}: "Excellent personalization. Seamlessly weaves in pertinent profile details; highly relevant and tailored guidance or conversation. Tone precisely matches the user's personality—empathetic, engaging, and concise. Avoids boilerplate and unnecessary disclaimers. Closely aligned with the user's likely preference as indicated by the reference."
\end{prompt}

\begin{prompt}{User Profile Generation Prompt}
\# Instruction\\
Generate a targeted user profile for \{\{ task\_description \}\} based on the provided user history data. This profile will be used to understand user behavior patterns specific to this task.\\
IMPORTANT: You must analyze HOW this user behaves and makes decisions relevant to this task, NOT list WHAT specific content they interact with. Do not output lists of topics, keywords, or content examples. Focus only on behavioral tendencies, preferences, and decision-making patterns relevant to the target task.\\
Focus on understanding patterns that inform task-specific user behavior:\\
1. Task-Relevant User Preferences:\\
    - Decision-making patterns and criteria relevant to the target task\\
    - Quality and content preferences that influence choices\\
    - Style and approach preferences in task-related activities\\
    - Consistency patterns in task-related decision making\\
2. Behavioral Patterns Related to Task Performance:\\
    - Interaction patterns and engagement styles relevant to the task\\
    - Response patterns to different types of content or options\\
    - Timing and frequency patterns in task-related activities\\
    - Adaptation patterns when encountering new or different scenarios\\
3. Personal Style and Voice Indicators:\\
    - Communication style patterns relevant to the task\\
    - Personal expression tendencies and voice characteristics\\
    - Authenticity markers and personal touch preferences\\
    - Consistency in personal style across different contexts\\
4. Context Awareness and Adaptation Patterns:\\
    - Awareness of audience, context, or requirements in task-related activities\\
    - Adaptation strategies for different scenarios within the task domain\\
    - Personalization approaches and individual preference integration\\
    - Balance between task requirements and personal style\\
\# User History Data: \{\{ user\_history \}\}\\
\# Output Format\\
Output the user profile strictly in plain text describing the user's behavioral patterns, preferences, and decision-making tendencies specifically relevant to \{\{ task\_description \}\}. Focus on patterns that predict how this user would approach and perform the target task.\\
Do NOT output:\\
- Lists of topics, keywords, or content they engage with\\
- Specific examples of content, products, or interactions\\
- Names of people, places, brands, or entities\\
- Content examples or subject matter details\\
DO output:\\
- Behavioral patterns and preferences relevant to the task\\
- Decision-making tendencies and criteria\\
- Personal style and approach patterns\\
- Task-specific interaction patterns\\
Derive insights strictly from the provided historical data. Do not include explanations, introductions, headings, bullet points, or any formatting structure.
\end{prompt}

\begin{prompt}{Task Description Prompt}
\texttt{LaMP-1 Citation Identification}: 'Academic citation recommendation: Identify relevant reference papers for researchers based on their publication titles and research focus areas.'\\

\texttt{LaMP-2N News Categorization}: 'News article categorization: Classify news articles into topical categories based on content, subject matter, and thematic focus.'\\

\texttt{LaMP-2M Movie Tagging}: 'Movie genre tagging: Analyze movie descriptions and assign appropriate genre tags based on content themes, narrative elements, and stylistic features.'\\

\texttt{LaMP-3 Product Rating}: 'Product review rating prediction: Analyze product reviews and predict the rating score based on sentiment, content quality, and expressed satisfaction levels.'\\

\texttt{LaMP-4 News Headline Generation}:  'News article headline generation: Create a concise and engaging headline for news articles based on their content and key themes.'\\

\texttt{LaMP-5 Scholarly Title Generation}: 'Academic title generation: Generate concise and descriptive titles for research papers based on abstracts, capturing the main research contribution and scope.'\\

\texttt{LaMP-7 Tweet Paraphrasing}: 'Tweet paraphrasing: Rewrite tweets in a personal style while maintaining the original meaning and adapting the tone and language to individual communication patterns.'\\

\texttt{LongLaMP-1 Abstract Generation}: 'Academic abstract generation: Create comprehensive abstracts for research papers based on titles and key research items, incorporating domain-specific knowledge and writing style.'\\

\texttt{LongLaMP-2 Topic Writing}: 'Topic-based content generation: Create personalized Reddit posts on given topics that reflect individual writing style, interests, and communication preferences.'\\

\texttt{LongLaMP-3 Review Writing}: 'Product review generation: Write detailed product reviews that reflect personal experiences, preferences, and writing style based on ratings and product features.'\\

\texttt{Empathetic Conversation}: 'Empathetic news commentary: Provide personal commentary and emotional reactions to news articles that reflect individual values, perspectives, and empathetic responses.'\\

\texttt{Personal Reddit}: 'Personal conversation response: Generate authentic and personalized responses to conversations that reflect individual personality, background, and communication style.'\\
\end{prompt}

\begin{prompt}{User Profile Embedding Prompt}
\# Instruction:\\
Extract and represent key personalization features that reflect the user's unique characteristics, preferences, and behavioral patterns from the provided user information. Focus on demographics, stated and implicit preferences, interaction history, behavioral trends, and other traits that can inform task-specific personalization.
Pay special attention to features relevant for the following task: \{\{ task\_description \}\}\\

\# User Information:\\
\{\{ user\_profile \}\}
\end{prompt}
\begin{prompt}{LaMP-1 Citation Identification Prompt}
\# Instruction: \\
    Identify the most relevant reference for the listed publication by the researcher. Select the reference paper that is most closely related to the researcher's work.\\

    \# Paper Title: \\
    \{paper\_title\}\\

    \# Options: \\
    \{options\_str\}\\

    \# Output Format: \\
    Just answer with [1] or [2] without explanation.\\
    
    \# Answer:\\
\end{prompt}

\begin{prompt}{LaMP-2N News Categorization Prompt}
\# Instruction: \\
    Which category does this article relate to among the following categories? Just answer with the category name without further explanation.\\

    \# Categories: \\
     travel, education, parents, style \& beauty, entertainment, food \& drink, science \& technology, business, sports, healthy living, women, politics, crime, culture \& arts, religion \\
    \# Article: \\
    \{article\}\\

    \# Output Format: \\
    Just answer the category without further explanation.\\
    
    \# Answer:\\
\end{prompt}

\begin{prompt}{LaMP-2M Movie Tagging Prompt}
\# Instruction: \\
    Which tag does this movie relate to among the following tags?\\

    \# Tags: \\
    sci-fi, based on a book, comedy, action, twist ending, dystopia, dark comedy, classic, psychology, fantasy, romance, thought-provoking, social commentary, violence, true story\\

    \# Description: \\
    \{description\}\\

    \# Output Format: \\
    Just answer with the tag name without further explanation.\\
    
    \# Answer:\\
\end{prompt}

\begin{prompt}{LaMP-3 Product Rating Prediction Prompt}
\# Instruction: \\
    What is the score of the following review on a scale of 1 to 5?\\

    \# Review: \\
    \{review\_text\}\\

    \# Output Format: \\
    Just answer with 1, 2, 3, 4, or 5 without further explanation.\\
    
    \# Answer:\\
\end{prompt}

\begin{prompt}{LaMP-4 News Headline Generation Prompt}
\# Instruction: \\
    Generate a headline for the following article.\\

    \# Article: \\
    \{article\}\\

    \# Output Format: \\
    Just answer the headline without further explanation.\\
    
    \# Answer:\\
\end{prompt}

\begin{prompt}{LaMP-5 Scholarly Title Generation Prompt}
\# Instruction:\\
    Generate a title for the following abstract of a paper.\\
    
    \# Abstract:\\
    \{abstract\}\\

    \# Output Format:\\
    Just answer the title without further explanation.\\
    
    \# Answer:\\
\end{prompt}

\begin{prompt}{LaMP-7 Tweet Paraphrasing Prompt}
\# Instruction:\\
    Paraphrase the following text into tweet without any explanation before or after it.\\
    
    \# Original Tweet:\\
    \{original\_tweet\}\\

    \# Output Format:\\
    Just answer the paraphrased tweet without further explanation.\\
    
    \# Answer:\\
\end{prompt}

\begin{prompt}{LongLaMP-1 Abstract Generation Prompt}
\# Instruction:\\
    Given the paper title and key items, generate an abstract for the paper.\\

    \# Paper Title:\\
    \{paper\_title\}\\

    \# Output Format:\\
    Only output the abstract. Do not include any explanation or formatting.\\
    
    \# Answer:\\
\end{prompt}

\begin{prompt}{LongLaMP-2 Topic Writing Prompt}
\# Instruction:\\
    Generate the content for a Reddit post based on the provided topic.\\

    \# Topic:\\
    \{topic\_prompt\}\\

    \# Output Format:\\
    Only output the reddit post. Do not include any explanation or formatting.\\
    
    \# Answer:\\
\end{prompt}    

\begin{prompt}{LongLaMP-3 Review Writing Prompt}
\# Instruction:\\
    Given the overall rating, product description, and review summary, generate the review text written by a reviewer.\\

    \# Overall Rating:\\
    \{rating\}\\

    \# Product Description:\\
    \{product\_description\}\\

    \# Review Summary:\\
    \{review\_summary\}\\

    \# Output Format:\\
    Only output the review text. Do not include any explanation or formatting.\\
    
    \# Answer:\\
\end{prompt}

\begin{prompt}{Emphathetic Conversation Prompt}
\# Article: \\
  \{article\}\\

  \# Instruction:\\ 
  After reading the news article, write an essay (300-800 characters) summarizing your thoughts and feelings about the article.\\
  
  \# Answer:\\
\end{prompt}

\section{Qualitative Examples}
We present qualitative examples in Table \ref{tab:lamp2example} and \ref{tab:lamp7example}. 

\begin{table*}[t]
\caption{User Profile and Generation Example from LaMP-2: Movie Tagging task.}
\begin{adjustbox}{max width=1\linewidth}
\begin{tabular}{p{2in}p{4in}}
    \toprule[1.5pt]
    \textbf{User Profile} & This user tends to categorize and tag movies with a strong emphasis on plot elements rather than stylistic elements. They show a consistent tendency to label movies with tags such as "dark comedy," "comedy," and "dystopia," indicating a preference for narratives that blend humor with darker themes or explore societal issues.  The user also demonstrates a clear sensitivity to specific movie genres like "sci-fi," "action," and "violence," often, indicating they enjoys them the user association and thematic elements and focusing user system tends to categorize movies movies movies based on these genres. They tend to recognize and tag hybrid genres, such as "dark comedy," showing an ability to identify films that blend multiple thematic elements.  In terms of genre recognition, the user shows a preference for mainstream tags over niche or artistic labels. They frequently use tags like "classic," "based on a book," and "fantasy," suggesting an appreciation for films that have cultural significance or are derived from literary works. The user exhibits a critical analysis depth in tag assignment, often choosing precise tags that reflect the complexity of the film's narrative or thematic content. They demonstrate a higher tagging confidence in genres they are knowledgeable about, such as "comedy" and "action."  Demographic influences play a role in the user's tag perception and assignment. They seem to prefer tags that align with their personal tastes, such as "comedy" and "dark comedy," while also recognizing the broader appeal of mainstream genres like "action" and "sci-fi." Overall, this user's tagging behavior indicates a focus on narrative structure and thematic complexity, with a strong preference for mainstream genres and a tendency to use precise, descriptive tags that reflect the film's content and style.\\
    \hline 
    \textbf{User Input} &  Peter Parker is an outcast high schooler abandoned by his parents as a boy, leaving him to be raised by his Uncle Ben and Aunt May. Like most teenagers, Peter is trying to figure out who he is and how he got to be the person he is today. As Peter discovers a mysterious briefcase that belonged to his father, he begins a quest to understand his parents' disappearance 2013 leading him directly to Oscorp and the lab of Dr. Curt Connors, his father's former partner. As Spider-Man is set on a collision course with Connors' alter ego, The Lizard, Peter will make life-altering choices to use his powers and shape his destiny to become a hero.\\
    \hline
    \textbf{Non-Personalized Output} & based on a book (\ding{55})\\
    \hline
    \textbf{RAG Output} & based on a book (\ding{55})\\
    \hline
    \textbf{PAG Output} & based on a book (\ding{55})\\
    \hline
    \textbf{OPPU} & comedy (\ding{55})\\
    \hline
    \textbf{\ourmethod{} Output} & action (\ding{51})\\
    \bottomrule[1.5pt]
\end{tabular}
\end{adjustbox}
\label{tab:lamp2example}
\end{table*}

\begin{table*}[t]
\caption{User Profile and Generation Example from LaMP-7: Tweet Paraphrasing task.}
\begin{adjustbox}{max width=1\linewidth}
\begin{tabular}{p{2in}p{4in}}
    \toprule[1.5pt]
    \textbf{User Profile} & This user tends to write in a casual and emotive style, often using informal language and expressing strong feelings. They frequently use contractions and colloquialisms, such as "totly" and "yup". Their tone can shift between affectionate ("awwwwww", "lol") and self-deprecating humor ("bitch", "like an abandoned child"). The user also shows a preference for sharing personal experiences and emotions, indicating a tendency to personalize content on personal on their on social media ands user. The user frequently uses abbreviations and emojis for emphasis, suggesting an expressive and engaging communication style. They tend to share personal details and express vulnerability, which could indicate a preference for authenticity and openness in their interactions. However, they also seem to value privacy, as seen in their reluctance to engage in conversations when feeling disturbed.In terms of paraphrasing, this user is likely to retain the core emotion and context of a tweet while personalizing the language. They may use more casual and emotionally charged words, incorporate personal anecdotes, and maintain a conversational tone. Their approach to paraphrasing suggests a focus on maintaining the original meaning through relatable and authentic expressions.\\
    \hline 
    \textbf{Original Tweet} & @msnye, I hope you pray for me because there's a chance I might end up at the Emergency Room tonight. \\
    \hline
    \textbf{Reference Answer} & @msnye u better.... Man o man pray for me cause I might end up @ the ER before the night is over. \\
    \hline
    \textbf{Non-Personalized Output} & @msnye Need your prayers, might head to ER tonight \\
    \hline
    \textbf{RAG Output} & @msnye I hope you're praying for me as I might head to the ER tonight.\\
    \hline 
    \textbf{PAG Output} & @msnye, I hope you're prayin' for me 'cuz I might hit the ER tonight \\
    \hline 
    \textbf{OPPU Output} & @msnye Need your prayers, thinking of heading to ER tonight. Fingers crossed!\\
    \hline
    \textbf{\ourmethod{} Output} & @msnye please pray for me tonight, I might be in the ER \\
    \bottomrule[1.5pt]
\end{tabular}
\end{adjustbox}
\label{tab:lamp7example}
\end{table*}
\end{document}